\documentclass[letterpaper, 10 pt, conference]{ieeeconf}
\IEEEoverridecommandlockouts                            
\usepackage[T1]{fontenc} 
\usepackage{enumerate}
\usepackage{amsfonts}
\usepackage[colorinlistoftodos]{todonotes}
\usepackage[font=small,skip=12pt]{caption}
\usepackage{float}
\usepackage{subcaption}

\usepackage{booktabs}
\makeatletter
\let\NAT@parse\undefined
\makeatother
\usepackage{hyperref}
\usepackage{verbatim}
\usepackage{comment}
\usepackage{amsmath}

\usepackage{mathtools}
\usepackage{dsfont}
\usepackage{tipa}
\usepackage{textcomp}
\DeclareMathAlphabet{\mathpzc}{OT1}{pzc}{m}{it}

\newcommand{\norm}[1]{\left\lVert#1\right\rVert}

\usepackage{graphicx}
\definecolor{ForestGreen}{rgb}{0.133, 0.545, 0.133}
\definecolor{CommentBlack}{rgb}{1.0, 0.0, 0.0}

\usepackage[ruled,vlined,linesnumbered]{algorithm2e}

\SetCommentSty{mycommfont}

\title{\LARGE \bf
Parameter Identification and Motion Control for Articulated Rigid Body Robots Using Differentiable Position-based Dynamics}

\author{Fei Liu$^{\dagger, 1}$, Mingen Li$^{\dagger, 1}$, Jingpei Lu$^1$, Entong Su$^1$ and Michael C. Yip$^1$ \IEEEmembership{Senior Member, IEEE}
\thanks{$\dagger$ These authors contributed equally. \protect\\ $^1$ Advanced Robotics and Controls Lab, University of California San Diego, La Jolla, CA 92093 USA. {\tt\small \{f4liu, mil025, jil360, ensu, yip\}@ucsd.edu}}
}

\begin{document}

\maketitle 
\thispagestyle{empty}
\pagestyle{empty}

\begin{abstract}
Simulation modeling of robots, objects, and environments is the backbone for all model-based control and learning. It is leveraged broadly across dynamic programming and model-predictive control, as well as data generation for imitation, transfer, and reinforcement learning. In addition to fidelity, key features of models in these control and learning contexts are speed, stability, and native differentiability. However, many popular simulation platforms for robotics today lack at least one of the features above. More recently, position-based dynamics (PBD) has become a very popular simulation tool for modeling complex scenes of rigid and non-rigid object interactions, due to its speed and stability, and is starting to gain significant interest in robotics for its potential use in model-based control and learning. Thus, in this paper, we present a mathematical formulation for coupling position-based dynamics (PBD) simulation and optimal robot design, model-based motion control and system identification. Our framework breaks down PBD definitions and derivations for various types of joint-based articulated rigid bodies. We present a back-propagation method with automatic differentiation, which can integrate both positional and angular geometric constraints. Our framework can critically provide the native gradient information and perform gradient-based optimization tasks. We also propose articulated joint model representations and simulation workflow for our differentiable framework. We demonstrate the capability of the framework in efficient optimal robot design, accurate trajectory torque estimation and supporting spring stiffness estimation, where we achieve minor errors. We also implement impedance control in real robots to demonstrate the potential of our differentiable framework in human-in-the-loop applications.


\end{abstract}

\section{Introduction}

\subsection{Optimization in Robotics}
In most control and planning problems, 
the standard approach to modeling robot kinematics and dynamics is by describing the robots as articulated rigid bodies that can be described efficiently by a set of partial differentiable equations derived from the Newton-Euler or Euler-Lagrange method \cite{Lynch_2017}. 
They provide a natural solution for solving static and dynamic solutions efficiently. However, with simulations and model predictions propagating in time, these models usually can only be solved using approximate integration schemes such as small timestep discretizations which can lead to inaccuracies in the robot states. In cases where explicit constraints must be satisfied, convergence can slow down dramatically and stability issues may arise. The problem becomes exacerbated when solving inverse problems, i.e., parameter identification, with explicit constraints given that there is no general inverse solution to the partial differentiable equations.  

Instead of directly solving the implicit robot motion equations, a prevalent strategy to forward and inverse problems for robot simulation involves relying on iteration through gradients. 
In most control or parameter identification problems, the gradients for the dynamic states, system parameters, and control policies should be provided to the controller for faster performance and better convergence properties. Regarding high degree-of-freedom (d.o.f.) robotic dynamics, the analytical derivations require dense computations that typically need to be approximated by finite differences. This also suffers from numerical rounding errors and slow convergence, and can generally be less stable. 

Another way to obtain gradients is through automatic differentiation (auto-diff). This programming approach builds a computation graph for backward differentiation by using the chain rule, but is limited by memory storage requirements for the graph  and the number of simulation steps used during
backpropagation. However, a major advantage that should not be minimized is that the computation graphs can be generated and computed in the backend, i.e., without user involvement. This has made autodifferentiability in simulation models a feature sought after by many scientists. With proper definitions, simulations that support this feature may be used with popular existing libraries for auto-diff, such as PyTorch \cite{Paszke_2019} and CppAD \cite{Bradley_2014}.


\begin{figure}
\vspace{2mm}
\centering
\includegraphics[width=0.99\linewidth]{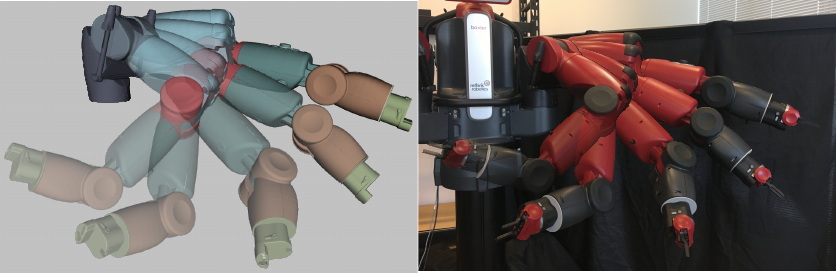}
\caption{Visualization of trajectory optimization with torque control. The right figure is an actual Baxter robot trajectory. The same motion is reproduced in simulation with torque control as in the left figure. Torque is optimized based on the differentiable PBD framework.
}
\label{fig:coverpage}
\vspace{-0.2in}
\end{figure}

Although many optimization tools have been developed, the state-of-art physical engines generally are not natively differentiable (e.g., Mujoco \cite{todorov_2012}, Bullet \cite{coumans_2021}), Gym \cite{brockman2016openai}, Nvidia Flex).
This is non-ideal for most applications involving optimization and learning, and with the heavy emphasis today in deep learning, there has been a push for differentiable physics and simulation \cite{Avila_2018, Hu_2020DiffTaichi, Qiao_2021, Werling_2021, Heiden_2021neuralsim, Tao_Du_2021, Zhiao_Huang_2021, Sirui_2021}. These approaches provided additional features such as ready-to-use gradients for optimization problems.

One of the most interesting developments for simulation and its potential for model-based control and learning is position-based dynamics (PBD) \cite{Macklin_2017}. 
Where the classical methods for a simulation involving Newton's second law  update force and velocity fields, and then finally positions are determined implicitly by numerical integration, position-based methods directly modify the positions for states updates. PBD approaches can thus solve kinematic and dynamic constraints by working with positions directly, and drastically improving both the speed, robustness, and accuracy of position-constrained simulations. It does this using non-linear projected Gauss-Seidel (NPGS) or Jacobi solvers, which are unconditionally stable.

While PBD's original development was towards complex rendering scenes for particle simulation, fluid, and soft bodies, rigid body simulation has been detailed through the formulation of position-based dynamics \cite{Muller_2020}, more recently, applications in robotics has been investigated. This includes deformable tissue modeling \cite{Fei_2021_ICRA, Yunhai_2021}, and blood fluid simulation \cite{Jingbin_2021_ICRA, Schenck_2018}. The obvious advantage of PBDs towards robot learning is that their speed, robustness, and native capability for defining constraints make it a scalable tool for eventually tackling complex scenarios and environments. However, to the best of our knowledge, only our previous works \cite{Jingbin_2021_ICRA, Schenck_2018} are differentiable, but these are not suitable for simulating articulated rigid body robots for lacking joint representations.

\subsection{Contributions}
In this work, we extend the approach proposed \cite{Muller_2020} to define a differentiable framework for robotics. Taking advantage of the introduction of positional and angular joint constraints, the robot system states will be represented by only several particles. 
We can then leverage auto-diff to differentiate through the forward dynamics computation of articulated rigid robots. We applied the differentiable solution for three typical robotics applications, i.e., optimal robot design, parameter identification, and motion control. Specifically, we summarized the paper's contributions below:
\begin{itemize}
    \item We propose the implementation of a position-based dynamics framework for articulated rigid body robots, involving joint constraints representations and updates,
    \item We formulate an optimization-based simulation workflow for differentiable framework by providing ready-to-use gradients,
    \item We demonstrate the utility of differentiable position-based physics (PBD) in optimal robotic design, parameter estimation, and motion control.
\end{itemize}

The novelty of our approach compared to existing differentiable works lies in introducing differentiability via computation graphs into a position-based framework. It maintains a sparse state representation using particles to represent articulated rigid robots. Our differentiable framework can provide both robust forward and backward position-based dynamics computation that works fluently with most optimization strategies in control and learning. 
We also implement our approach both on simulation and the 7-d.o.f Baxter robot arms (Rethink Robotics,  GmBH) to show the real-world effectiveness, as a result shown in Fig. \ref{fig:coverpage}.

\section{Related Works}

In this section, we cover recent work in differentiable rigid physical simulation. Benefiting from the new capabilities enabled by differentiability, several interesting robotic problems can be formulated into optimization problems with ready-to-use gradients. We then review three typical robotic applications that have been investigated in the literature. These application areas will also be targeted in our subsequent proposed method for differentiable rigid body simulation using PBDs.

\subsection{Differentiable Rigid Physical Simulation}

An end-to-end differentiable simulator that can be embedded into deep neural networks is proposed in \cite{Avila_2018}. The rigid body dynamics are simulated with a linear complementarity problem (LCP) by considering contact and friction constraints. However, only 2D rigid bodies in Cartesian coordinates are validated, insufficient for high d.o.f articulated robots. Meanwhile, the presence and absence of collision are non-differentiable. By LCP formulation, a more efficient differentiable simulation of articulated bodies is presented \cite{Qiao_2021}. The adjoint method derives the analytical gradients through spatial algebra faster than auto-diff tools and requires less memory. However, the effectiveness is only tested on simulated robots. Similarly, the analytical gradients are also provided \cite{Werling_2021} through LCP solution. The elastic collision gradients are approximated in continuous time to support complex contact geometry. This method is also integrated into an existed physics engine DART \cite{Jeongseok_2018} for performance verification. However, the proposed method is not able to differentiate through the robot geometric properties, such as link length and inertia. Thus, it will be limited for parameter estimation applications. 
Neural networks have been deployed into an augmented differentiable rigid-body engine in \cite{Heiden_2021neuralsim}. This hybrid simulator shows the ability to learn complex dynamics from real data by replacing the Quadratic programming (QP) solver inside a model-predictive controller with neural layers. Some other differentiable rigid simulated are introduced in \cite{Degrave_2019, Millard_2020automatic, Lutter_2020differentiable}. However, they are mostly applied to learning-based tasks with adaptation to neural networks. Apart from the rigid body dynamics modeling, the multi-body differentiable framework is proposed in \cite{Geilinger_2020, Hu_2020DiffTaichi, Murthy_2021gradsim}. They are applicable to both rigid and deformable objects within a unified method. The former one analytically computes derivatives via sensitivity analysis, while the latter two use source-code transformation for gradients backpropagation. 
Although several existing works have been investigated for rigid body simulation. However, only a few of them have considered a real field robot for dynamics problems, such as motion control. Meanwhile, the states spaces representation still mainly falls into traditionally Euler-Lagrange variables, which requires intensive calculation of analytical gradients.

a) \textit{Optimal Robotic Design}: The design and building of a robot is cost- and time-consuming which normally requires many iterations for the final version. It is challenging to ensure the design parameters without incorporating motion control constraints (such as d.o.f, work-space, etc.). In literature, the parameterization methods based on primitive shapes have been proposed using gradient-based \cite{Sehoon_2021} and gradient-free \cite{Xinlei_2021_ICRA} approaches for optimizations of design. However, these primitive shapes are limited by simple geometrical features. Recently, the differentiable simulation have been applied to optimizing the control scheme and the robot design jointly, in \cite{Spielberg_2019, Jie_Xu_2021, Pingchuan_2021, Tao_Du_2021}, namely the co-design of robots.

Differentiability within the end-to-end framework can result in better performance with a significantly fewer number of iterations, compared to gradient-free approaches. More importantly, more complex morphology can be optimized by providing analytical gradients for data efficiency.
For instance, an interesting robotic application for automated routing and controlling of muscle fibers has been researched in \cite{Maloisel_2021}. 
The differentiability of deformation gradient modeled by a hyperelastic material simulation, can be obtained by a moving least-squares formulation. It paved a way for data-driven actuator modeling and control over generalized artificial muscles design. Similar works regarding active-actuator modeling have also been implemented in \cite{Pingchuan_2021, Sehee_2019}. In our following work, we will consider the co-design of a robot arm by optimizing kinematics and dynamics variables, as well as control parameters simultaneously.



b) \textit{Dynamical Motion Control}: Optimization plays as a key method solving dynamical motion control and trajectory planning for robots. 
Differentiable works in \cite{Bern_2019, Qiao_2021_Multibody, Heiden_2021} can natively support for model-based motion control, such as dynamic locomotion, contact handling etc. In these works, the sensitivity analysis (also known as adjoint method) approach were employed to calculate the analytical derivatives of motion trajectories with respect to control polices. Thus, the optimal motion control can be generated from a high-level forward cost function directly. In \cite{Qiao_2021_Multibody}, the authors compared their methods to the state-of-art reinforcement learning approaches.
The advantages of differentiability were also highlighted in comparison with other gradient-based and gradient-free methods in \cite{Heiden_2021}. In the following sections, dynamically compliant motion control (i.e., impedance modulation) is formulated as an optimization problem and demonstrated.


c) \textit{Parameter Identification} : 
In recent works, the reinforcement learning approach suffers from the reality gap between simulation and real life \cite{Miguel_2021}. To reduce the reality gap, a first step beyond using idealized models is to have the parameterized mathematical model fit to measurements of the real world \cite{Chiuso_2019, Ramos_2019}. This is basically solving the inverse problem for model-based control and learning. Recent works with differentiable simulation lead to a new path for real-to-sim or sim-to-real applications. Therefore, the differentiability can be easily applied for system estimation and parameter identification such as in \cite{Lutter_2021, Lidec_2021, Wang_Kun_2020_pmlr, Carolyn_2020, Jiajun_2015}. The dynamics of the system can be differentiable regarding parameters by defining a loss function over sensors or even images. In this way, the data efficiency allows a cheaper transfer of the simulated model to the real world. In this work, we also demonstrate this capability by optimizing parameters from collected real data.

\section{Differentiable PBD Simulation}
This section presents the forward dynamics for articulated rigid body robots using position-based dynamics (PBD). Then, we introduce the differentiability of the method for robotic applications.


\subsection{Articulated Rigid Body Position-based Dynamics}

Position-based dynamics (PBD) has been a popular approach in recent years for fast simulation of particle-based dynamics. A detailed review for common objects such as cloth, deformable bodies, and fluid can be found in \cite{Macklin_2017}. These works fundamentally show the modeling of deformable materials as particles in Cartesian space with relationships between nearest neighbors. While most geometric constraints are defined only as a function of the positional information of particles and their adjacent members, orientation plays important role in simulations of objects such as rigid bodies. The approaches using shape matching \cite{Muller_2005_ShapeMatching}, and oriented particles \cite{Muller_2011_OrientedParticle} have been proposed as an extension of PBD to handle orientation and angular for arbitrary numbers and arrangements of particles. However, these methods still need a dense particle cloud to represent a single rigid body, which is not computation efficient.
In \cite{Deul_2016}, positional constraints are introduced to handle rigid-body dynamics. Only one particle is needed for the representation of each link. However, how to model articulated rigid bodies using this sparse representation is not defined.

To define a rigid body which can both translate and rotate in space, the particle representation per link (i.e., rigid body) is extended with orientation information to model joint kinematic constraints. It should be noted that each link of the articulated robot connected by joints can be represented as a single particle with both positional and angular constraints.
Then, the PBD for both soft and rigid articulated bodies was solved jointly as a constraint-based optimization problem.
As a contribution, we follow the classical position-based dynamics pipeline to present the derivation of articulated constraints. To clarify, a list of parameters used is in Table \ref{tab:parameters_list}. 

\begin{figure*}[!htbp]
\vspace{2mm}
\centering
\includegraphics[width=1.0\textwidth]{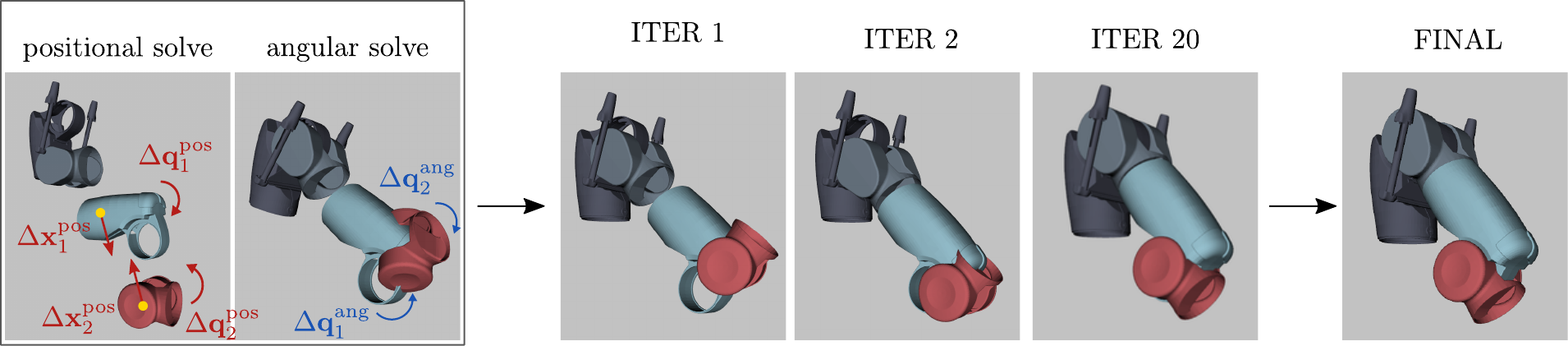}
\caption{The iterations of translational and rotational constraints correction updates using a projected Gauss-Seidel solver of PBD.
}
\label{fig:updating_solvers_steps}
\end{figure*}

In the simulation, a rigid body link that within the articulated system is represented by (i) a particle (located at the center of mass, COM) with Cartesian coordinates $\mathbf{x} \in \mathbb{R}^3$ and (ii) quaternion $\mathbf{q} \in \mathbb{R}^4$ representing the orientation of the body, mass $m$ and moments of inertia $\mathbf{I} \in \mathbb{R}^3$, as well as  vectors $\prescript{}{l}{\mathbf{t}_{i}}, \prescript{}{l}{\mathbf{r}_{i}} \in \mathbb{R}^3$ pointing from the COM towards 
its and the next hinge joint position. To join two links together via a hinge joint, three different constraints need to be considered for each joint, i.e. positional, angular and angle limit constraints. The non-linear projected Gauss-Seidel solver will update the translational and rotational correction iteratively. For a more comprehensive demonstration, the solving of constraints convergence is shown in Fig. \ref{fig:updating_solvers_steps}. The robot links are detached from each other for the initial state.

\begingroup
\setlength{\tabcolsep}{10pt} 
\renewcommand{\arraystretch}{1.2} 
\begin{table}[!tbp]
    \caption{NOMENCLATURE
    }
    \centering
    \begin{tabular}{p{3.0cm} p{4.4cm}}
    \toprule[1pt]
    $i \in \left[1,2, \cdots, N\right]$ & links represented by particles\\
    $\mathbf{x}_i \in \mathbb{R}^3$, $\mathbf{v}_i  \in \mathbb{R}^3$  & COM position/velocity \\
    $\mathbf{q}_i \in \mathbf{SO}(3)$  &  link quaternion \\
    $\mathds{R}(*) \in \mathbf{SO}(3)$  &  rotation matrix \\
    $\boldsymbol{\phi}_i \in \mathrm{Im}\left(\mathbb{H}\right)$  &  link's Rodrigues rotation vector\\
    $\boldsymbol{\omega}_i \in \mathbb{R}^3$   &   angular velocity \\
    $m_i \in \mathbb{R}$, $\mathbf{I}_i \in \mathbb{R}^{3\times3}$  &  mass/inertia tensor \\
    $\mathbf{f}_{ext} \in \mathbb{R}^3$, $\boldsymbol{\tau}_{ext} \in \mathbb{R}^3$  & external force/torque\\
    $\prescript{}{l}{\mathbf{t}_{i}} \in \mathbb{R}^{3} $  &  local joint vectors from COM of link $i$ connected to link $i-1$ \\
    $\prescript{}{l}{\mathbf{r}_{i}} \in \mathbb{R}^{3} $  & local joint vectors from COM of link $i$ connected to link $i+1$ \\
    $\left[
    \prescript{}{l}{\mathbf{a}_{i}^x}, \prescript{}{l}{\mathbf{a}_{i}^y}, \prescript{}{l}{\mathbf{a}_{i}^z}
    \right] 
    \in \mathbb{R}^{3\times3}$
    & local joint axes of link $i$ connected to link $i-1$ \\
    $\left[
    \prescript{}{l}{\mathbf{b}_{i}^x}, \prescript{}{l}{\mathbf{b}_{i}^y}, \prescript{}{l}{\mathbf{b}_{i}^z}
    \right] 
    \in \mathbb{R}^{3\times3} $
    & local joint axes of link $i$ connected to link $i+1$ \\
    $\lambda$ &  Lagrangian multiplier\\
    $\mathbf{C}^{\mathcal{P}}, \mathbf{c}^{\mathcal{P}}$ & positional constraints \\
    $\mathbf{C}^{\mathcal{A}}, \mathbf{c}^{\mathcal{A}}$ & angular constraints \\
    \bottomrule[1pt]
    \end{tabular}
    \label{tab:parameters_list} 
\end{table}
\endgroup

\begin{figure}[h!]
\vspace{2mm}
\centering
\includegraphics[width=0.99\linewidth]{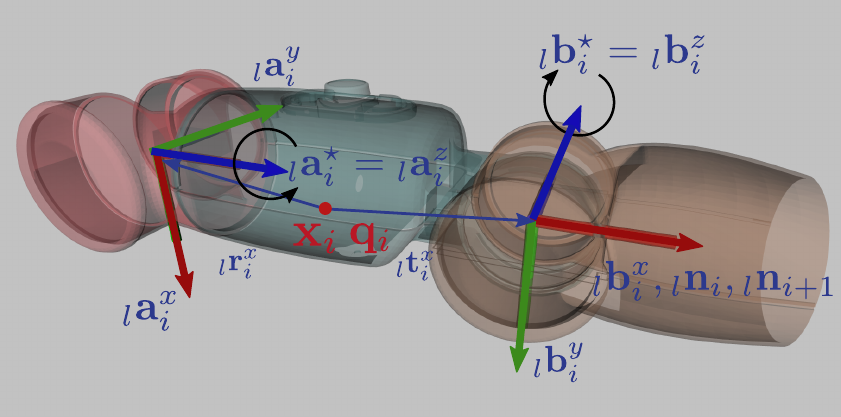}
\caption{Baxter arm and variables visualization. Variables in red are in global frame and variables in blue are in link i local frame. Rotational axis $\prescript{}{l}{\mathbf{a}^{\star}_{i}}$ and  $\prescript{}{l}{\mathbf{b}^{\star}_{i}}$ for attached joints are annotated with rotation mark. In the figure, link axis $_{l}\mathbf{n}_i$, $_{l}\mathbf{n}_{i+1}$ are the same because current joint angle is 0 rad.
}
\label{fig:variable_fig}
\end{figure}

\subsubsection{Positional Constraint}
We begin with two connected links, each represented by particles located at their respective centers of mass, i.e., $\mathbf{x}_{i}, \mathbf{x}_{i+1}$, 
with the quaternion representing orientation denoted as $\mathbf{q}_{i}, \mathbf{q}_{i+1}$ respectively. 
The positional constraint aims to solve the correction terms at their center of masses that ensure that the particle distances relative to the hinge are constant.
A generalized distance constraint should be satisfied between each particle pair resulting in $3N$ constraints, i.e.,  
$\mathbf{C}^{\mathcal{P}}(\mathbf{x}, \mathbf{q})=\left\{ \mathbf{c}_i^{\mathcal{P}}(\mathbf{x}_{i}, \mathbf{x}_{i+1}, \mathbf{q}_{i}, \mathbf{q}_{i+1})~|~i \in \left[1,2, \cdots, N\right] \right\}$, 
where
\begin{equation}
\label{eq:joint_dis_const}
\begin{split}
    \mathbf{c}_i^{\mathcal{P}}(\mathbf{x}_{i}, \mathbf{x}_{i+1}, \mathbf{q}_{i}, \mathbf{q}_{i+1}) =  & \left[ \mathbf{x}_{i+1} + \prescript{l}{w}{\mathds{R}(\mathbf{q}_{i+1})} \prescript{}{l}{\mathbf{t}_{i+1}} \right] 
    \\
    & - \left[ \mathbf{x}_{i} + \prescript{l}{w}{\mathds{R}(\mathbf{q}_{i})} \prescript{}{l}{\mathbf{r}_{i}} \right]
\end{split}
\end{equation}
is in $\mathbb{R}^3$ and where $\prescript{}{l}{\mathbf{t}_{i+1}}$ and $\prescript{}{l}{\mathbf{r}_{i}}$ are local position vector to the hinge relative to the COM and $\prescript{l}{w}{\mathds{R}(\cdot)}$ is the rotation matrix represented by rotation vector, from local frame to world frame. 


The constraints may be satisfied via gradient steps where the steps are defined via Taylor expansion,
\begin{equation}
\label{eq:linearization}
{\small
\begin{split}
    & \mathbf{c}_i^{\mathcal{P}}(\mathbf{x}_i + \Delta\mathbf{x}_i^\mathcal{P}, \mathbf{q}_i \oplus \Delta\boldsymbol{\phi}_i^\mathcal{P})  \\
    & \approx  
    \mathbf{c}_i^{\mathcal{P}}(\mathbf{x}_i,\mathbf{q}_i) + \nabla_{\mathbf{x}}\mathbf{c}_i^\mathcal{P}(\mathbf{x}_i,\mathbf{q}_i) \Delta\mathbf{x}_i^\mathcal{P} + \nabla_{\boldsymbol{\phi}}\mathbf{c}_i^\mathcal{P}(\mathbf{x}_i,\mathbf{q}_i) \Delta\boldsymbol{\phi_i}^\mathcal{P} \\
    & =  0
\end{split}
}
\end{equation}
with $\Delta \mathbf{x}_i^\mathcal{P} \in \mathbb{R}^3$ describing a small step in position of the link COM and $\Delta \mathbf{\phi}_i^\mathcal{P} \in \mathbb{R}^3$ representing a small rotation the be applied to the orientation vector of the link, and
$\nabla_{\mathbf{x}}\mathbf{c}_i^\mathcal{P}(\mathbf{x}_i,\mathbf{q}_i)$, $\nabla_{\boldsymbol{\phi}}\mathbf{c}_i^\mathcal{P}(\mathbf{x}_i,\mathbf{q}_i) \in \mathbb{R}^{1\times 3}$ representing the gradient of the constraint with respect to the position and rotation variables. Given this, the positional and orientation updates are defined as
\begin{equation}
\label{eq:delta_positional_vector}
\begin{split}
    &  \Delta\mathbf{x_i}^\mathcal{P}  =  \mathbf{W}_{\mathbf{x}_i}^\mathcal{P}\nabla^\top_\mathbf{x} \mathbf{c}_i^\mathcal{P}(\mathbf{x}_i, \mathbf{q}_i){\lambda}^\mathcal{P}_{\mathbf{x}_i} 
    \\
    &  \Delta\boldsymbol{\phi}^\mathcal{P}  = \mathbf{W}^\mathcal{P}_{\boldsymbol{\phi}_i} \nabla^\top_{\boldsymbol{\phi}} \mathbf{c}_i^\mathcal{P}(\mathbf{x}_i, \mathbf{q}_i){\lambda}^\mathcal{P}_{\mathbf{\phi}_i} 
\end{split}
\end{equation}
Weight terms $\mathbf{W}_{\mathbf{x}_i}, \mathbf{W}_{\mathbf{\phi}_i}$ are introduced in order to conserve linear and angular momentum by limiting the correction direction to being parallel to the gradient. Ideally, the correction should be proportional to the pair-wise generalized inverse masses  and rotational inertia, i.e., 
\begin{equation}
\mathbf{W}_{\mathbf{x}_i} = m_i^{-1}\cdot diag\{1,1,1\} \text{~~ ~~}
\mathbf{W}_{\mathbf{\phi}_i} = \mathbf{I}_i^{-1}\\
\end{equation}


It should be noticed the concatenated vector $\boldsymbol{\Lambda}^\mathcal{P}$ is only one scalar value regarding each constraint index $i$. Since each constraint $\mathbf{c}_i^{\mathcal{P}}$ is a three-dimensional vector and its elements are isotropically weighted, we introduce $l^2$-norm (i.e., $\left\lVert \cdot \right\rVert$) for simplification. Consequently, by taking Eq. \ref{eq:delta_positional_vector} into Eq. \ref{eq:linearization}, we can get the following relationship regarding the Lagrange multipliers:
\begin{equation}
\label{eq:Lagrange_multiplier}
    \lambda^\mathcal{P}_{i} = - \left(\sum_{
    \substack{\pi \in  \{\mathbf{x}_{i}, \mathbf{x}_{i+1},\\ 
    \Delta\boldsymbol{\phi}_{i}, \Delta\boldsymbol{\phi}_{i+1}\}}
    } \nabla_{\pi} \left\lVert\mathbf{c}_i^{\mathcal{P}}\right\rVert \mathbf{W}_{\pi}^{\mathcal{P}} \nabla^\top_{\pi} \left\lVert\mathbf{c}_i^\mathcal{P}\right\rVert \right)^{-1} \left\lVert\mathbf{c}_i^{\mathcal{P}}\right\rVert
\end{equation}
and our simplified positional update rules will be,
\begin{equation}
\label{eq:positional_updates_with_lambda}
\begin{split}
    & \Delta\mathbf{x}^\mathcal{P}_{k \in \{i, i+1\}}  =  \mathbf{W}_{\mathbf{x}_{k}}^\mathcal{P} \nabla^\top_{\mathbf{x}_{k}} \left\lVert\mathbf{c}_i^\mathcal{P}\right\rVert \lambda^\mathcal{P}_i \\
    & \Delta\boldsymbol{\phi}^\mathcal{P}_{k \in \{i, i+1\}}  = \mathbf{W}_{\Delta\boldsymbol{\phi}_{k}}^\mathcal{P} \nabla^\top_{\Delta\boldsymbol{\phi}_{k}} \left\lVert\mathbf{c}_i^\mathcal{P}\right\rVert  \lambda^\mathcal{P}_i 
\end{split}
\end{equation}

Adapting to the positional distance constraint $\mathbf{C}^{\mathcal{P}}$ in Eq. \ref{eq:joint_dis_const},
each constraint gradient can be calculated as, 
\begin{equation}
\begin{split}
    & \nabla^\top_{\mathbf{x}_{i}} \left\lVert\mathbf{c}_i^\mathcal{P}\right\rVert = - \mathbf{1}_{3 \times 3} \mathbf{n}_i^{\mathcal{P}} \\
    & \nabla^\top_{\mathbf{x}_{i+1}} \left\lVert\mathbf{c}_i^\mathcal{P}\right\rVert =  \mathbf{1}_{3 \times 3} \mathbf{n}_i^{\mathcal{P}} \\
    & \nabla^\top_{\Delta\boldsymbol{\phi}_{i}} \left\lVert\mathbf{c}_i^\mathcal{P}\right\rVert = - \frac{\partial 
    \prescript{l}{w}{\mathds{R}(\mathbf{q}_{i})} \prescript{}{l}{\mathbf{r}_{i}}
    }{
    \partial \Delta\boldsymbol{\phi}_{i}
    }  \mathbf{n}_i^{\mathcal{P}} 
    = 
    \left[
    \prescript{l}{w}{\mathds{R}(\mathbf{q}_{i})} \prescript{}{l}{\mathbf{r}_{i}}
    \right]^{\times} 
    \mathbf{n}_i^{\mathcal{P}}  \\
    & \nabla^\top_{\Delta\boldsymbol{\phi}_{i+1}} \left\lVert\mathbf{c}_i^\mathcal{P}\right\rVert = \frac{\partial 
    \prescript{l}{w}{\mathds{R}(\mathbf{q}_{i+1})} \prescript{}{l}{\mathbf{t}_{i+1}}
    }{
    \partial \Delta\boldsymbol{\phi}_{i+1}
    }  \mathbf{n}_i^{\mathcal{P}}
    \\
    & ~~~~~~~~~~~~~~~  =
    - \left[
    \prescript{l}{w}{\mathds{R}(\mathbf{q}_{i+1})} \prescript{}{l}{\mathbf{t}_{i+1}}
    \right]^{\times}
     \mathbf{n}_i^{\mathcal{P}} \\
\end{split}
\end{equation}
where $\displaystyle \mathbf{n}_i^{\mathcal{P}} = \mathbf{c}_i^{\mathcal{P}}/\left\lVert\mathbf{c}_i^{\mathcal{P}}\right\rVert$, and the gradients regarding rotation vector is derived from Lie Group according to the left perturbation scheme, and $[*]^{\times}$ is the skew-symmetric representation.

Then, by following the derivative of generalized inverse masses with impulse based rigid body solvers in \cite{Muller_2020}, it leads to,
\begin{equation}
\begin{split}
    & \nabla_{\mathbf{x}_{i}} 
    \left\lVert\mathbf{c}_i^\mathcal{P}\right\rVert \mathbf{W}_{\mathbf{x}_{i}}^{\mathcal{P}} 
    \left(\nabla_{\mathbf{x}_{i}} \left\lVert\mathbf{c}_i^\mathcal{P}\right\rVert\right)^\top  
        = m_i^{-1}\\
    & \nabla_{\mathbf{x}_{i+1}} 
    \left\lVert\mathbf{c}_i^\mathcal{P}\right\rVert  \mathbf{W}_{\mathbf{x}_{i+1}}^{\mathcal{P}} 
    \left(\nabla_{\mathbf{x}_{i}} \left\lVert\mathbf{c}_i^\mathcal{P}\right\rVert \right)^\top  
        = m_{i+1}^{-1}\\
    & \nabla_{\Delta\boldsymbol{\phi}_{i}} \left\lVert\mathbf{c}_i^\mathcal{P}\right\rVert \mathbf{W}_{\Delta\boldsymbol{\phi}_{i}}^{\mathcal{P}} 
    \left(\nabla_{\Delta\boldsymbol{\phi}_{i}} \left\lVert\mathbf{c}_i^\mathcal{P}\right\rVert\right)^\top  \\
        & ~~ = \left( 
        \left[\prescript{l}{w}{\mathds{R}(\boldsymbol{\phi}_{i})} \prescript{}{l}{\mathbf{r}_{i}}\right]^{\times}
        \mathbf{n}_i^{\mathcal{P}} \right)^\top 
        \mathbf{I}^{-1}_{i} 
        \left(
        \left[\prescript{l}{w}{\mathds{R}(\boldsymbol{\phi}_{i})} \prescript{}{l}{\mathbf{r}_{i}}\right]^{\times}
        \mathbf{n}_i^{\mathcal{P}} \right)
         \\
    & \nabla_{\Delta\boldsymbol{\phi}_{i+1}} \left\lVert\mathbf{c}_i^\mathcal{P}\right\rVert \mathbf{W}_{\Delta\boldsymbol{\phi}_{i+1}}^{\mathcal{P}} 
    \left(\nabla_{\Delta\boldsymbol{\phi}_{i+1}} \left\lVert\mathbf{c}_i^\mathcal{P}\right\rVert\right)^\top  \\
        & ~~  = \left(
        -\left[\prescript{l}{w}{\mathds{R}(\boldsymbol{\phi}_{i+1})} \prescript{}{l}{\mathbf{t}_{i+1}}\right]^{\times}
        \mathbf{n}_i^{\mathcal{P}} \right)^\top 
        \mathbf{I}^{-1}_{i+1} \\
        & ~~~~~~~~~~~~~~~~~~~~~~~ \left(
        - \left[\prescript{l}{w}{\mathds{R}(\boldsymbol{\phi}_{i+1})} \prescript{}{l}{\mathbf{t}_{i+1}}\right]^{\times}
        \mathbf{n}_i^{\mathcal{P}} \right)
        \\
\end{split}
\end{equation}

Combining the above equations with Eq. \ref{eq:Lagrange_multiplier} for Lagrange multiplier $\lambda_i$, one can quickly obtain the following implementation for distance constraints of positional updates in Eq. \ref{eq:positional_updates_with_lambda},
\begin{equation}
\label{eq:positional_updates_delta_x}
\begin{split}
    & \Delta \mathbf{x}^{\mathcal{P}}_{i} =  - m_{i}^{-1} \cdot \lambda^{\mathcal{P}}_i 
    \cdot \mathbf{n}_i^{\mathcal{P}} \\
    & \Delta \mathbf{x}^{\mathcal{P}}_{i+1} = m_{i+1}^{-1} \cdot \lambda^{\mathcal{P}}_{i} 
    \cdot \mathbf{n}_i^{\mathcal{P}} \\
    & \Delta \boldsymbol{\phi}^{\mathcal{P}}_{i} =  \mathbf{I}_{i}^{-1} 
    \left( 
    \left[\prescript{l}{w}{\mathds{R}(\boldsymbol{\phi}_{i})} \prescript{}{l}{\mathbf{r}_{i}}
    \right]^{\times} 
    \cdot \mathbf{n}_i^{\mathcal{P}}
    \right) \cdot \lambda^{\mathcal{P}}_i \\
    & \Delta \boldsymbol{\phi}^{\mathcal{P}}_{i+1} = - \mathbf{I}_{i+1}^{-1} 
    \left(
    \left[\prescript{l}{w}{\mathds{R}(\boldsymbol{\phi}_{i+1})} \prescript{}{l}{\mathbf{t}_{i+1}}
    \right]^{\times}
    \cdot \mathbf{n}_i^{\mathcal{P}}
    \right) \cdot  \lambda^{\mathcal{P}}_i\\
\end{split}
\end{equation}

The rotation represented with quaternions will also be corrected simultaneously. Following the small value changes of the rotation vector for Lie group in \cite{Sola_2017quaternion}, the updates of quaternions $\Delta \mathbf{q}^{\mathcal{P}}_i$, $\Delta \mathbf{q}^{\mathcal{P}}_{i+1}$ for alignment of orientation can be approximated with Euler method as,
\begin{equation}
\label{eq:positional_updates_delta_q}
\begin{split}
    & \Delta \mathbf{q}^{\mathcal{P}}_{i} = \frac{1}{2} \left[ 0, \Delta \boldsymbol{\phi}^{\mathcal{P}}_{i} 
    \right] \otimes \mathbf{q}_{i}\\
    &  \Delta \mathbf{q}^{\mathcal{P}}_{i+1} = \frac{1}{2} \left[ 0, \Delta \boldsymbol{\phi}^{\mathcal{P}}_{i+1}
    \right] \otimes \mathbf{q}_{i+1}\\
\end{split}
\end{equation}
where $\otimes$ indicates the multiplication of two quaternions. It should be noticed that the above inertia tensors depend on the rotation center. Thus, they must be updated in the global world frame after every constraint projection solving.

\subsubsection{Angular Constraint}
Hinge joint angular constraint aims at aligning the rotational axes of two connected links (i.e., $i$ and $i+1$) attached to the same hinge joint, regarding their local joint position vector from COM, as $\prescript{}{l}{\mathbf{r}_{i}}$ and $\prescript{}{l}{\mathbf{t}_{i+1}}$ respectively. Let $\prescript{}{l}{\mathbf{b}^{\star}_{i}}\in\mathbb{R}^3$ and $\prescript{}{l}{\mathbf{a}^{\star}_{i+1}}\in\mathbb{R}^3$ denote the normalized rotational axis vector in local frame for link $i$ and link $i+1$. Then, a generalized angular constraint should be satisfied by $\mathbf{C}^{\mathcal{A}}(\mathbf{q})=\left\{ \mathbf{c}_i^{\mathcal{A}}(\mathbf{q}_{i}, \mathbf{q}_{i+1})~|~i \in \left[1,2, \cdots, N\right] \right\}$. Thus, the angular constraint should be defined by,
\begin{equation} 
\begin{split}
    \mathbf{c}_i^{\mathcal{A}}(\boldsymbol{\phi}_{i}, \boldsymbol{\phi}_{i+1}) 
    & = \left[ \prescript{l}{w}{\mathds{R}(\boldsymbol{\phi}_{i})} \prescript{}{l}{\mathbf{b}^{\star}_{i}} \right]
    \times
    \left[ \prescript{l}{w}{\mathds{R}(\boldsymbol{\phi}_{i+1})} \prescript{}{l}{\mathbf{a}^{\star}_{i+1}} \right] \\
    & = \Delta \boldsymbol{\phi}_{i} = - \Delta \boldsymbol{\phi}_{i+1}
\end{split}
\label{eq:joint_angular_const}
\end{equation}
with angular positional updates,
\begin{equation}
\begin{split}
    & \Delta \boldsymbol{\phi}^\mathcal{A}_{k \in \{i, i+1\}}  = \mathbf{W}_{\Delta\boldsymbol{\phi}_{k}}^\mathcal{A} \left(\nabla_{\Delta\boldsymbol{\phi}_{k}} \mathbf{c}_i^\mathcal{A}\right)^\top \lambda^\mathcal{A}_i 
\end{split}
\end{equation}

Then, we can follow the same derivation for positional constraints as above, the Lagrange multiplier for angular constraints can be obtained by,
\begin{equation}
\label{eq:Lagrange_multiplier_angular}
\begin{split}
    \lambda^\mathcal{A}_{i} = - \left( 
    \sum_{ \substack{\pi \in  \{\Delta\boldsymbol{\phi}_{i}, \Delta\boldsymbol{\phi}_{i+1}\}} }
    \nabla_{\pi} \mathbf{c}_i^\mathcal{A} \mathbf{W}_{\pi}^{\mathcal{A}} \left(\nabla_{\pi} \mathbf{c}_i^\mathcal{A} \right)^\top \right)^{-1} \cdot \mathbf{c}_i^\mathcal{A}
\end{split}
\end{equation}


Similarly, each angular constraint gradient can be calculated through Lie group derivation as, 
\begin{equation}
\begin{split}
    & \nabla^\top_{\Delta\boldsymbol{\phi}_{i}} \left\lVert\mathbf{c}_i^\mathcal{A}\right\rVert = \mathbf{n}_i^{\mathcal{A}} \\
    & \nabla^\top_{\Delta\boldsymbol{\phi}_{i+1}} \left\lVert\mathbf{c}_i^\mathcal{A}\right\rVert = - \mathbf{n}_i^{\mathcal{A}}
\end{split}
\end{equation}

\noindent where $\displaystyle \mathbf{n}_i^{\mathcal{A}} = \mathbf{c}_i^{\mathcal{A}}/\left\lVert\mathbf{c}_i^{\mathcal{A}}\right\rVert$. By multiplying generalized inverse masses, we obtain
\begin{equation}
\begin{split}
    & \nabla_{\boldsymbol{\phi}_{i}} \left\lVert\mathbf{c}_i^{\mathcal{A}}\right\rVert \mathbf{W}_{\boldsymbol{\phi}_{i}}^{\mathcal{A}} 
    \left(\nabla_{\boldsymbol{\phi}_{i}} \left\lVert\mathbf{c}_i^{\mathcal{A}}\right\rVert\right)^\top 
        = \left( \mathbf{n}_i^{\mathcal{A}} \right)^\top 
          \mathbf{I}^{-1}_{i} 
          \mathbf{n}_i^{\mathcal{A}} \\
    & \nabla_{\boldsymbol{\phi}_{i+1}} \left\lVert\mathbf{c}_i^{\mathcal{A}}\right\rVert \mathbf{W}_{\boldsymbol{\phi}_{i+1}}^{\mathcal{A}} 
    \left(\nabla_{\boldsymbol{\phi}_{i+1}} \left\lVert\mathbf{c}_i^{\mathcal{A}}\right\rVert \right)^\top 
        = \left( \mathbf{n}_i^{\mathcal{A}} \right)^\top 
          \mathbf{I}^{-1}_{i} 
          \mathbf{n}_i^{\mathcal{A}} \\
\end{split}
\end{equation}
which leads to, 
\begin{equation}
\label{eq:angular_updates_delta_q}
\begin{split}
    & \Delta\boldsymbol{\phi}^\mathcal{A}_{i}  = \mathbf{I}^{-1}_{i} \mathbf{n}_i^{\mathcal{A}} \lambda^\mathcal{A}_i \\
    & \Delta\boldsymbol{\phi}^\mathcal{A}_{i+1}  = - \mathbf{I}^{-1}_{i+1} \mathbf{n}_i^{\mathcal{A}} \lambda^\mathcal{A}_i 
\end{split}
\end{equation}
The updates of quanternion $\Delta \mathbf{q}^{\mathcal{A}}_i$, $\Delta \mathbf{q}^{\mathcal{A}}_{i+1}$ is,
\begin{equation}
\label{eq:positional_updates_delta_q}
\begin{split}
    & \Delta \mathbf{q}^{\mathcal{A}}_{i} = \frac{1}{2} \left[ 0, \Delta \boldsymbol{\phi}^{\mathcal{A}}_{i} 
    \right] \otimes \mathbf{q}_{i}\\
    &  \Delta \mathbf{q}^{\mathcal{A}}_{i+1} = \frac{1}{2} \left[ 0, \Delta \boldsymbol{\phi}^{\mathcal{A}}_{i+1}
    \right] \otimes \mathbf{q}_{i+1}\\
\end{split}
\end{equation}


\subsection{Differentiability}
To make our forward PBD simulation easily differentiable and accessible for learning problems, we specifically cater the method to automatic differentiation (auto-diff) frameworks such as PyTorch \cite{PyTorch}. Fig. \ref{fig:flow_diagram_MPC_torque_estimation} gives an overview of the differentiable simulation. The gradients will be tracked and stored by a sequence of elementary mathematical operations. 
The parameter and system states can be optimized by the backpropagation of a computation graph with a defined loss function.
A practical challenge is that the number of time steps of forward simulation that might be leveraged for backpropagating losses is limited by the memory required for automatic differentiation.
In practice, this problem can be addressed by wrapping up certain functions, computing intermediate results and storing these intermediate checkpoints for later use.

\section{Parameter Identification and Model-Predictive Control}
The ultimate objective of the differentiable simulation is to perform sim-to-real applications, where there is a mismatch between the simulation and the real world and solving an inverse problem can help reduce this gap. These scenarios present themselves in problems of parameter identification, as well as certain model-predictive control problems. Below we lay out the methods for parameter identification and model-predictive control that include parameter estimation via backpropagating error through the entire control/simulation chain (which includes our differentiable model). 


Consider the problem of parameter estimation, where we aim to infer the parameters $\boldsymbol{\alpha}$ (i.e., inertia $\mathbf{I}$ etc.) and system states $\boldsymbol{\xi}$ (i.e., link represented by PBD particle position $\mathbf{x}$, velocity $\mathbf{v}$, orientation $\mathbf{q}$, and angular velocity $\boldsymbol{\omega}$), or control signals $\mathbf{u}$ (i.e., input torque $\boldsymbol{\tau}$) using position-based constraints. The optimization problem can be written implicitly as,

\begin{equation}
\begin{split}
        \boldsymbol{\alpha}^*, \boldsymbol{\xi}^*, \mathbf{u}^*
    & = \operatorname*{argmin}_{\boldsymbol{\alpha}, \boldsymbol{\xi}, \mathbf{u}} \mathcal{L} \left( \mathbf{x}, \mathbf{v}, \mathbf{q}, \boldsymbol{\omega} \,|\, \boldsymbol{\alpha}, \mathbf{u} \right) \\
        \operatorname*{s.t.} ~~
    & \mathbf{C}^{\mathcal{P}}(\mathbf{x}, \mathbf{q}) = 0 \\
    & \mathbf{C}^{\mathcal{A}}(\mathbf{q}) = 0
\end{split}
\end{equation}
Where $\mathbf{C}^{\mathcal{P}}$ and $\mathbf{C}^{\mathcal{A}}$ are the position-based constraints, $\mathcal{L}$ is the loss function generated from the representation of forward PBD simulation.


\begin{algorithm}[!htbp]
    \caption{Differentiable Articulated Robot Framework}
    \label{alg:differentiable_framework}
    \SetKwInOut{Input}{Input}
    \SetKwInOut{Output}{Output}
    \Input{Current time-step link representation with particle position $\mathbf{x}^{t}$, velocity $\mathbf{v}^{t}$, quaternion $\mathbf{q}^{t}$, angular velocity $\boldsymbol{\omega}^{t}$, control torque $\boldsymbol{\tau}^{t}$, gravity $\mathbf{g}$, local inertia tensor $\prescript{}{l}{\mathbf{I}^0}$ at time-step 0 etc.
    }
    \Output{Gradients of loss function $\mathcal{L}$ regarding system parameters $\boldsymbol{\alpha}^{t}$, PBD system states $\boldsymbol{\xi}^t$, or control input $\mathbf{u}^{t}$}

    \tcp{Initialize corresponding gradients variables}
    $\boldsymbol{\alpha}^{t} \leftarrow \lbrace \mathrm{inertia}~\prescript{}{l}{\mathbf{I}^0}, \cdots \cdots \rbrace $ \\
    $\mathbf{u}^{t} \leftarrow \boldsymbol{\tau}^{t}$ \\
    $\boldsymbol{\xi}^t \leftarrow \mathbf{x}^{t}, \mathbf{v}^{t}, \mathbf{q}^{t}, \boldsymbol{\omega}^{t}$
    
    
        \tcp{Position states Euler prediction}
        $\mathbf{v}^{t+1} \leftarrow \mathbf{v}^{t} + \Delta{t} \cdot \mathbf{g} $ \\
        $\mathbf{x}^{t+1} \leftarrow \mathbf{x}^{t} + \Delta{t} \cdot \mathbf{v}^{t}$\\
        
        \tcp{Update inertia tensor in global frame}
        $\mathbf{I}^{t} =  \prescript{l}{w}{\mathds{R}^\top(\mathbf{q}^{t})}  \cdot \prescript{}{l}{\mathbf{I}^0} \cdot \prescript{l}{w}{\mathds{R}(\mathbf{q}^{t})} $ \\
        
        \tcp{Angular states Euler prediction}
        $\small \boldsymbol{\omega}^{t+1} \leftarrow \boldsymbol{\omega}^{t} + \Delta{t} \cdot (\mathbf{I}^t)^{-1} \left(\mathbf{u}^{t} + \boldsymbol{\omega}^{t} \times \left(\mathbf{I}^t \boldsymbol{\omega}^{t}\right)\right)$ \\
        $\mathbf{q}^{t+1} \leftarrow \mathbf{q}^{t} + \frac{1}{2} \Delta{t} \cdot[0, \boldsymbol{\omega}^{t+1}]\mathbf{q}^{t}$ \\
        $ \mathbf{q}^{t+1} \leftarrow \mathbf{q}^{t+1} / \left\lVert \mathbf{q}^{t+1} \right\rVert$\\
        
        ~ \\
        \tcp{Constraints solving loop}
        \For{$k$ iterations}{
            \tcp{Apply positional constraints using Eq. \ref{eq:positional_updates_delta_x} and Eq. \ref{eq:positional_updates_delta_q}}
            $\Delta{\mathbf{x}^{\mathcal{P}}}, \Delta{\mathbf{q}^{\mathcal{P}}} \leftarrow \mathtt{solvePositional}\left(\mathbf{C}^{\mathcal{P}}\left(\mathbf{x}^{t+1}, \mathbf{q}^{t+1}\right) = 0\right)$ \\
            $\mathbf{x}^{t+1} \leftarrow \mathbf{x}^{t+1} + \Delta{\mathbf{x}^{\mathcal{P}}}$\\
            $\mathbf{q}^{t+1} \leftarrow \mathbf{q}^{t+1} + \Delta{\mathbf{q}^{\mathcal{P}}}$ \\
            $\mathbf{q}^{t+1} \leftarrow \mathbf{q}^{t+1} / \left\lVert \mathbf{q}^{t+1} \right\rVert$ \\
            
            \tcp{Apply angular constraints using Eq. \ref{eq:angular_updates_delta_q} }
            $\Delta{\mathbf{q}^{\mathcal{A}}} \leftarrow \mathtt{solveAngular}\left(\mathbf{C}^{\mathcal{A}}\left(\mathbf{q}^{t+1}\right) = 0\right)$\\
            $\mathbf{q}^{t+1} \leftarrow \mathbf{q}^{t+1} + \Delta{\mathbf{q}^{\mathcal{A}}}$ \\
            $\mathbf{q}^{t+1} \leftarrow \mathbf{q}^{t+1} + \Delta{\mathbf{q}^{\mathcal{P}}}$ \\
            $\mathbf{q}^{t+1} \leftarrow \mathbf{q}^{t+1} / \left\lVert \mathbf{q}^{t+1} \right\rVert$ \\
        } 
        
        ~ \\
        \tcp{Position states Euler integration}
        $\mathbf{v}^{t+1} \leftarrow \left( \mathbf{x}^{t+1} - \mathbf{x}^{t}\right)/\Delta{t}$ \\

        \tcp{Angular states Euler integration}
        $\Delta{\mathbf{q}}^{t+1} \leftarrow \mathbf{q}^{t+1} (\mathbf{q}^{t})^{-1}$ \\
        $\boldsymbol{\omega}^{t+1} \leftarrow 2 \cdot \Delta{\mathbf{q}}^{t+1} / \Delta{t} $ \\
        \uIf{$\mathfrak{R}(\Delta{\mathbf{q}}^{t+1}) >= 0$}{
            $\boldsymbol{\omega}^{t+1} \leftarrow \boldsymbol{\omega}^{t+1}$ 
        }
        \Else{
            $\boldsymbol{\omega}^{t+1} \leftarrow -\boldsymbol{\omega}^{t+1}$
        }
        
        
        

        \tcp{Obtain the loss function}
        $\mathcal{L} \leftarrow \mathcal{L} \left( \mathbf{x}^{t+1}, \mathbf{v}^{t+1}, \mathbf{q}^{t+1}, \boldsymbol{\omega}^{t+1} \right)$\\

        \tcp{Calculate the gradients}
	    $\displaystyle{\frac{\partial \mathcal{L}}{\partial \boldsymbol{\alpha}^{t}}, \frac{\partial \mathcal{L}}{\partial \boldsymbol{\xi}^t}, \frac{\partial \mathcal{L}}{\partial \mathbf{u}^t} }=\texttt{autodiff}(\pi, \boldsymbol{\xi})$\\
	    
	    \vspace{3mm}
        \Return{$\displaystyle{\frac{\partial \mathcal{L}}{\partial \boldsymbol{\alpha}^{t}}, \frac{\partial \mathcal{L}}{\partial \boldsymbol{\xi}^t}, \frac{\partial \mathcal{L}}{\partial \mathbf{u}^t} }$}\\
\end{algorithm}

\subsection{Parameter Identification}
For dynamical systems, a traditional Euler-Lagrange formulation requires generalized non-conservative force as input to the system $\boldsymbol{\tau} \in \mathbb{R}^{N}$, which is described by,
\begin{equation}
\label{eq:eulerlagrange}
\begin{split}
        \frac{\mathrm{d}}{\mathrm{dt}} \frac{\partial L(\mathbf{p}, \dot{\mathbf{p}})}{\partial \dot{\mathbf{p}}} - \frac{\partial L(\mathbf{p}, \dot{\mathbf{p}})}{\partial \mathbf{p}} = \boldsymbol{\tau}
\end{split}
\end{equation}
with the Lagrangian value defined by,
\begin{equation}
\begin{split}
        L = T - V
\end{split}
\end{equation}
where $T$ and $V$ are the kinetic and potential energy respectively, $\mathbf{p}$ and $\dot{\mathbf{p}}$ are the generalized coordinates of a robotic system. For joint-space values, we can assume $\left[\mathbf{p}, \dot{\mathbf{p}}, \ddot{\mathbf{p}}\right]$ are the joint level position, velocity, and acceleration respectively, and $\mathbf{\tau}$ are torques applied to the joints.
Werling and Lutter \cite{Werling_2021, Lutter_2020} present the Recursive-Newton-Euler (RNE) algorithm for the kinetic and potential models for differentiability.

In sim-to-real application, it is convenient to transform the PBD states to joint values with a kinematic mapping. In this way, the control strategy can be described in joint-space which converts naturally to motor trajectories after accounting for gearing and transmission couplings (if present). We denote function $\boldsymbol{f}$ as the kinematic mapping from PBD states $\boldsymbol{\xi} \in \left[ \boldsymbol{\mathbf{x}}, \boldsymbol{\mathbf{v}}, \boldsymbol{\mathbf{q}}, \boldsymbol{\omega} \right]$ to the generalized joint states $\left[ \mathbf{p}, \dot{\mathbf{p}} \right]$ of the simulated robot, i.e.,
\begin{equation}
\begin{split}
        \pi_{\substack{ \in \{ \mathbf{p}, \dot{\mathbf{p}} \}}}
    & = \boldsymbol{f}_{ \substack{\pi} } \left( \boldsymbol{\mathbf{x}}, \boldsymbol{\mathbf{v}}, \boldsymbol{\mathbf{q}}, \boldsymbol{\omega} \right) \\
    & = \boldsymbol{f}_{ \substack{\pi} } \left( \bigcup_{ \substack{\boldsymbol{\xi} \in  \{ \boldsymbol{\mathbf{x}}, \boldsymbol{\mathbf{v}}, \boldsymbol{\mathbf{q}}, \boldsymbol{\omega} \}} } \boldsymbol{\xi} \right)
\end{split}
\end{equation}

\noindent Then without loosing generality, a loss function for parameter identification involving joint values can be explicitly defined as,
\begin{equation}
\label{eq:para_ident_optimal}
\begin{split}
        \boldsymbol{\alpha}^* 
    & = \operatorname*{argmin}_{\boldsymbol{\alpha}} \mathcal{L} \left( \mathbf{p}, \dot{\mathbf{p}} \right) \\
    & = \operatorname*{argmin}_{\boldsymbol{\alpha}} \sum_{ \substack{\hat{\pi} \in  \{
    \hat{\mathbf{p}}, \hat{\dot{\mathbf{p}}} \}} } \left\lVert \hat{\pi} - \pi \right\rVert^2 \\
    & = \operatorname*{argmin}_{\boldsymbol{\alpha}} \sum_{ \substack{\hat{\pi} \in  \{ \hat{\mathbf{p}}, \hat{\dot{\mathbf{p}}} \}} } \left\lVert \hat{\pi} - \boldsymbol{f}_{\substack{\pi}} \left( \bigcup  \boldsymbol{\xi} \right) \right\rVert^2
\end{split}
\end{equation}
where $\hat{\pi}$ is the corresponding observed target data $\hat{(\cdot)}$ used in the loss definition, and is task-dependent and coming from sensor data. 
For example, it can be the joint-level trajectory way-points over time 
$\left( \mathbf{p}^0, \mathbf{p}^1, \cdots, \mathbf{p}^t \right)$,
or joint velocity over time 
$\left( \dot{\mathbf{p}}^0, \dot{\mathbf{p}}^1, \cdots, \dot{\mathbf{p}}^t \right)$.

In order to solve the problem defined in Eq. \ref{eq:para_ident_optimal}, the optimization can be performed by iterative gradient stems calculated via chain-rule as,
\begin{equation}
\label{eq:para_ident_chain_rule}
\begin{split}
    \frac{\partial \mathcal{L}}{\partial \boldsymbol{\alpha}} 
    & =  \sum_{\substack{\pi \in  \{ \mathbf{p}, \dot{\mathbf{p}} \}}} \frac{\partial \mathcal{L}}{\partial \pi} \frac{ \partial \pi }{ \boldsymbol{\partial \alpha} }  \\
    & =  \sum_{\substack{\pi \in  \{ \mathbf{p}, \dot{\mathbf{p}} \} }} \sum_{\substack{\boldsymbol{\xi} \in  \{ \boldsymbol{\mathbf{x}}, \boldsymbol{\mathbf{v}}, \boldsymbol{\mathbf{q}}, \boldsymbol{\omega} \}}} 
    \frac{\partial \mathcal{L}}{\partial \pi} \frac{\partial \pi}{\partial \boldsymbol{\xi}}
    \frac{\partial \boldsymbol{\xi}}{ \partial \boldsymbol{\alpha} }
\end{split}
\end{equation}

\noindent The first partial derivative (i.e., $\frac{\partial \mathcal{L}}{\partial \pi}$) can be straightforward to calculate as long as the loss function is explicitly expressed. Normally it will be a combination or a single joint-level trajectory value, i.e., $\pi$. The second partial derivative (i.e., $\frac{\partial \pi}{\partial \boldsymbol{\xi}}$) depends on the forward kinematic mapping from PBD states to joint states. In our case, we proposed the kinematic mapping via iterative linear transforms (shown algorithmically in Algorithm \ref{alg:joint_angle_conver}). 

One might notice that for Eq. \ref{eq:para_ident_chain_rule}, the joint states of joint velocity may be considered. In this case, we don't directly get the loss function regarding $\dot{\mathbf{p}}$, but instead we process it by transferring the observed data (i.e., $\hat{\dot{\mathbf{p}}}$) back into PBD state space, that is, 
\begin{equation}
\begin{split}
    \prescript{}{w}{\hat{\boldsymbol{\omega}}}_{w,n}^t = & \sum_{i=1}^{n} \prescript{l}{w}{\mathds{R}(\mathbf{q}_{i}^{t})}
    {}_{l}{\mathbf{b}_{i}^{\star}} \hat{\dot{\mathbf{p}}}_{i}^{t}\\
\end{split}
\label{equ:joint_velcity_to_PBD_States}
\end{equation}
In this way, the loss function can be calculated using $\boldsymbol{\omega}$ from PBD directly, which will only rely on gradient of $\frac{\partial \boldsymbol{\xi}}{ \boldsymbol{\partial \alpha} }$ which has already been obtained from Algorithm \ref{alg:differentiable_framework}. Therefore, the final chain-rule gradient can be reformulated as,  
\begin{equation}
\label{eq:para_ident_chain_rule_reformulated}
\begin{split}
    \frac{\partial \mathcal{L}(\mathbf{p}, \boldsymbol{\omega})}{\partial \boldsymbol{\alpha}} 
    & = \sum_{i} \sum_{j} 
    \frac{\partial \mathcal{L}}{\partial \boldsymbol{\mathbf{p}}_i} \frac{\partial \boldsymbol{\mathbf{p}}_i}{\partial \mathbf{q}_j}
    \frac{\partial \mathbf{q}_j}{ \boldsymbol{\partial \alpha} } + \sum_{k} \frac{\partial \boldsymbol{\omega}_k}{ \partial \boldsymbol{\alpha} }
\end{split}
\end{equation}
where $i$, $j$, $k$ are the index of each joint. All the involved gradients can be obtained from Algorithm \ref{alg:differentiable_framework} and Algorithm \ref{alg:joint_angle_conver}.

\begin{algorithm}[!htb]
    \caption{The gradient of kinematic mapping from PBD states to joint states (joint angle)}
    \label{alg:joint_angle_conver}
    \SetKwInOut{Input}{Input}
    \SetKwInOut{Output}{Output}
    \Input{The link axis $\prescript{}{l}{\mathbf{n}_{i}}$ and $\prescript{}{l}{\mathbf{n}_{i+1}}$ with shared joint rotation axis ${}_{l}{\mathbf{b}_{i}^{\star}}$ (or ${}_{l}{\mathbf{a}_{i+1}^{\star}}$) in local frame as shown in Fig. \ref{fig:variable_fig}, quaternions $\mathbf{q}_{i}$ and $\mathbf{q}_{i+1}$ for link $i$ and link $i+1$ respectively
    }
    \Output{Gradients of loss function $\displaystyle \frac{\partial \pi}{\partial \boldsymbol{\xi}}$}

        \tcp{transfer link axes from local to world frame}
		$\prescript{}{w}{\mathbf{n}_{i}} = \mathds{R}(\mathbf{q}_i^t) \prescript{}{l}{\mathbf{n}_{i}} $ \\
		$\prescript{}{w}{\mathbf{n}_{i+1}} = \mathds{R}(\mathbf{q}_{i+1}^t) \prescript{}{l}{\mathbf{n}_{i+1}} $ \\
		\tcp{transfer shared joint rotation axis from local to world frame}
		${}_{w}{\mathbf{b}_{i}^{\star}} = \mathds{R}(\mathbf{q}_i^t) {}_{l}{\mathbf{b}_{i}^{\star}}$ \\
		
		~ \\
		\tcp{introduce joint limit constraint within $[-\pi,\pi]$}
		$\mathbf{p}_{i} = \arcsin(\prescript{}{w}{\mathbf{n}_{i}}
		   \times \prescript{}{w}{\mathbf{n}_{i+1}}) \cdot 
		    {}_{w}{\mathbf{b}_{i}^{\star}}$ \\
	    \If{$(\prescript{}{w}{\mathbf{n}_{i}}
	    \cdot \prescript{}{w}{\mathbf{n}_{i+1}}) < 0$}{$\mathbf{p}_{i} = \pi - \mathbf{p}_{i}$}
	    \If{$\mathbf{p}_{i}> \pi$} {$\mathbf{p}_{i} = \mathbf{p}_{i}-2\pi$  \tcp{$\mathbf{p}_{i} \in [-\pi,\pi]$}} 
	    \If{$\mathbf{p}_{i} < \pi$}{$\mathbf{p}_{i} = \mathbf{p}_{i}+2\pi$ \tcp{$\mathbf{p}_{i} \in [-\pi,\pi]$}} 
	  
	    \tcp{Obtain the gradients values}
        $\pi \leftarrow \mathbf{p}_{i}$\\
        $\boldsymbol{\xi} \leftarrow \mathbf{q}_{i}, \mathbf{q}_{i+1}$\\
	  
	    \tcp{Calculate the gradients using auto-diff}
	    $\frac{\partial \pi}{\partial \boldsymbol{\xi}}=\texttt{autodiff}(\pi, \boldsymbol{\xi})$\\
	    \Return{$\displaystyle \frac{\partial \pi}{\partial \boldsymbol{\xi}}$}
\end{algorithm}

\subsection{Torque Estimation From Trajectory}

Similar to the parameter identification problem, we now will define a loss function for torque estimation as 
\begin{equation}
\begin{split}
    \boldsymbol{u}^* 
    & = \operatorname*{argmin}_{\boldsymbol{u}} \mathcal{L} \left( \mathbf{p}, \boldsymbol{\omega} \right) \\
    & = \operatorname*{argmin}_{\boldsymbol{u}} \frac{1}{K} \sum_{t} \sum_{i} \lambda_1 \left\lVert \boldsymbol{p}^t_i - \hat{\boldsymbol{p}^t_i} \right\rVert ^2 + \\
    & ~ \lambda_2 \norm{ \boldsymbol{\omega}_{i}^{t} - \prescript{}{w}{\hat{\boldsymbol{\omega}}}_{w, i}^t }^2 + 
    \lambda_3 \norm{ {\boldsymbol{u}}_{i}^t - {\boldsymbol{u}}_{i}^{t-1} }^2
\end{split}
\label{equ:lossfunctiontorque}
\end{equation}
with factor $\lambda_1,\lambda_2,\lambda_3$ for weight adjustment and $1/K$ being the average factor for MSE, the loss function includes not only mean square error terms for joint angles, angular velocity to simulate true trajectory, but also counting of torque difference for minimizing torque change to avoid overshooting. Indices $i$ and $t$ enumerate the joints and time steps, respectively.

Similar to Eq. \ref{eq:para_ident_chain_rule_reformulated}, the chain-rule gradient can be formulated by,
\begin{equation}
\label{eq:para_ident_chain_rule_reformulated}
\begin{split}
    \frac{\partial \mathcal{L}(\mathbf{p}, \boldsymbol{\omega})}{\partial \boldsymbol{u}} 
    & = \sum_{t} \sum_{i} \sum_{j} 
    \frac{\partial \mathcal{L}}{\partial \boldsymbol{\mathbf{p}}^t_i} \frac{\partial \boldsymbol{\mathbf{p}}^t_i}{\partial \mathbf{q}^t_j}
    \frac{\partial \mathbf{q}^t_j}{ \boldsymbol{\partial u^t} } + \sum_{t} \sum_{k} \frac{\partial \boldsymbol{\omega}^t_k}{\partial \boldsymbol{u}^t} \\
    & ~ + \mathtt{constant}
\end{split} 
\end{equation}
Compared to Eq. \ref{eq:para_ident_chain_rule_reformulated}, the difference for this chain-rule involves all time-steps for torque estimation in order to obtain a continuous control input. 

\begin{figure}
\vspace{2mm}
\centering
 \includegraphics[width=0.99\linewidth]{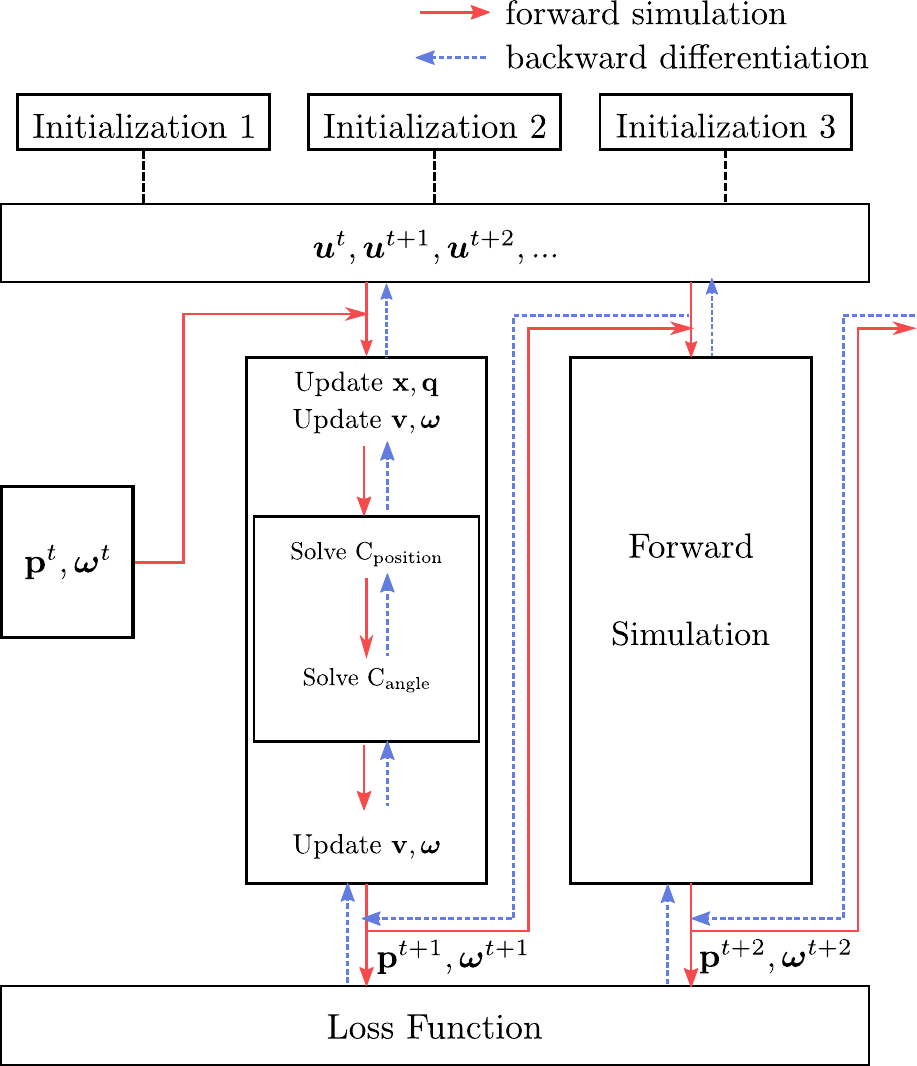}
\caption{Flow diagram of differentiable simulation and torque estimation with model-predictive control (MPC) pattern. To learn the motor torque, forward simulation in the red line is executed and then backpropagation through the blue line.
}
\label{fig:flow_diagram_MPC_torque_estimation}
\vspace{-0.2in}
\end{figure}

We are also able to set up the torque estimation in a model-predictive control (MPC) pattern. For example, we will simultaneously initialize and estimate three (or more) torque control inputs (i.e., $\boldsymbol{u}^{t}$, $\boldsymbol{u}^{t+1}, \boldsymbol{u}^{t+2}$, $\cdots$) at time frame $t$. We will get the summarized loss function for all predictive states (i.e., $\boldsymbol{\xi}^{t+1}$, $\boldsymbol{\xi}^{t+2}$, $\boldsymbol{\xi}^{t+3}$, $\cdots$). We can then solve for the finite horizon control commands using a concatenated loss, where forward simulation and backward calculation are concatenated one by one, as shown in Fig. \ref{fig:flow_diagram_MPC_torque_estimation}.



\subsection{Impedance Control in Joint Space}
When the robot encounters environmental contact that can be unsafe to the robot or the environment, impedance control has often been a method of choice for imposing a safer dynamic profile. Here, we propose joint impedance control laws comprising proportional–derivative (PD) controllers and model-based components for contact force prediction as in Eq. \ref{eq:impedance_equation}. 
\begin{equation}
\begin{split}
     \boldsymbol{u} = & \underbrace{\boldsymbol{D} (\boldsymbol{\dot{q}}_d - \boldsymbol{\dot{q}}) + \boldsymbol{K} (\boldsymbol{q}_d-\boldsymbol{q})}_{\boldsymbol{\tau}_{pd}} + \\
    &  \underbrace{\boldsymbol{M}(\boldsymbol{q})\boldsymbol{\ddot{q}}_d + 
    \boldsymbol{S} (\boldsymbol{q},\boldsymbol{\dot{q}})\boldsymbol{\dot{q}} + \boldsymbol{g}(\boldsymbol{q})}_{\boldsymbol{\tau}_{net}}
\end{split}
\label{eq:impedance_equation}
\end{equation}

The PD control terms $\boldsymbol{\tau}_{pd}$ includes damping term $\boldsymbol{D}$ and stiffness term $\boldsymbol{K}$. Dynamics models could be used for calculating net torque $\boldsymbol{\tau}_{net}$ includes inertial, Coriolis, and gravitational elements, which would be predicted based on optimization in our differentiable simulation framework with a given initial state and a final state.




\section{Experiment and Result}
In this section, we implement our differentiable framework and conduct gradient-based optimization tasks, including robot design, parameter estimation, torque estimation, and impedance control. Inherently all these problems involve the use of gradient information in some way, with some solving forward while other solving inverse problems for robot control and model learning. We show that the method is capable of being used in all of these scenarios successfully.

\subsection{Experimental Setup}
Our differentiable simulation framework uses a Jacobi solver with Pytorch. Constraints are solved in 30 iterations to guarantee convergence and gravity is set to 9.8 $m/s^2$ with time interval $\Delta t=0.01s$.
The Baxter robot in simulation consists of  7 Baxter links and 7 hinge joints corresponding to their shoulder, elbow, and wrist ($s_0, s_1, e_0, e_1, w_0, w_1, w_2$). The first link is attached to a base link with infinite mass so that the base is still. To configure the Baxter left arm, parameters in the Unified Robot Description Format (URDF) file provided in the ROS package \texttt{baxter\_common}\footnote{\url{https://github.com/RethinkRobotics/baxter_common}}
are utilized. While the file provides inertia tensor, mass, etc.,
the positional and rotational information in local frames needs transformation to the global frame.\\

\begin{figure}
\vspace{2mm}
\centering
\includegraphics[width=0.99\linewidth]{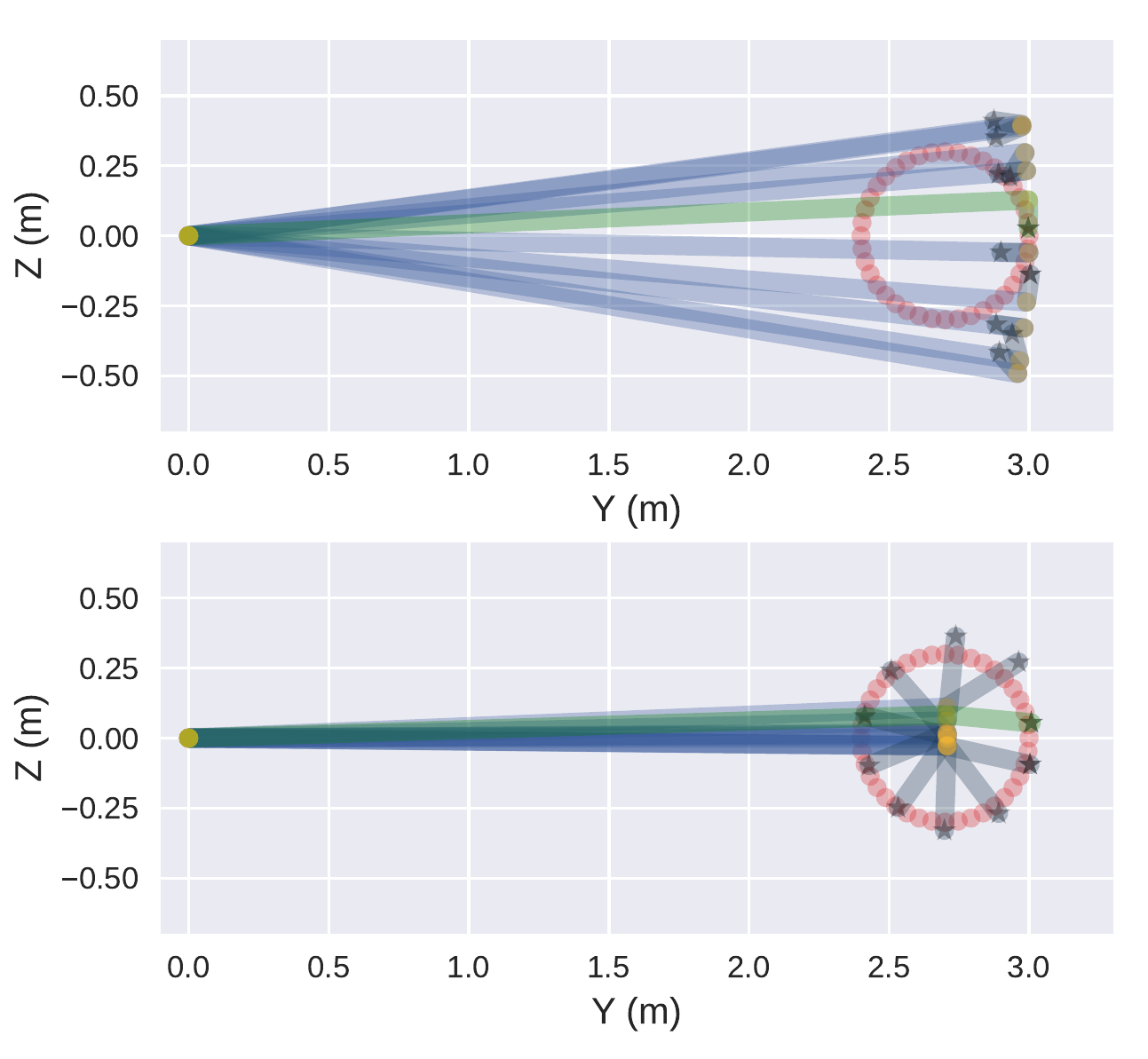}
\caption{Top plot shows the robot arm before optimizing the length of links. The bottom plot shows the robot arm with optimized length of links. The green arm is the initial position. The desired trajectory is red. The trajectory of the robot's end-effector using the controller is black starred. The joint points are yellow.
} 
\label{fig:pendulumplot}
\vspace{-0.2in}
\end{figure}

\begin{figure}
\vspace{2mm}
\centering
\includegraphics[width=0.99\linewidth]{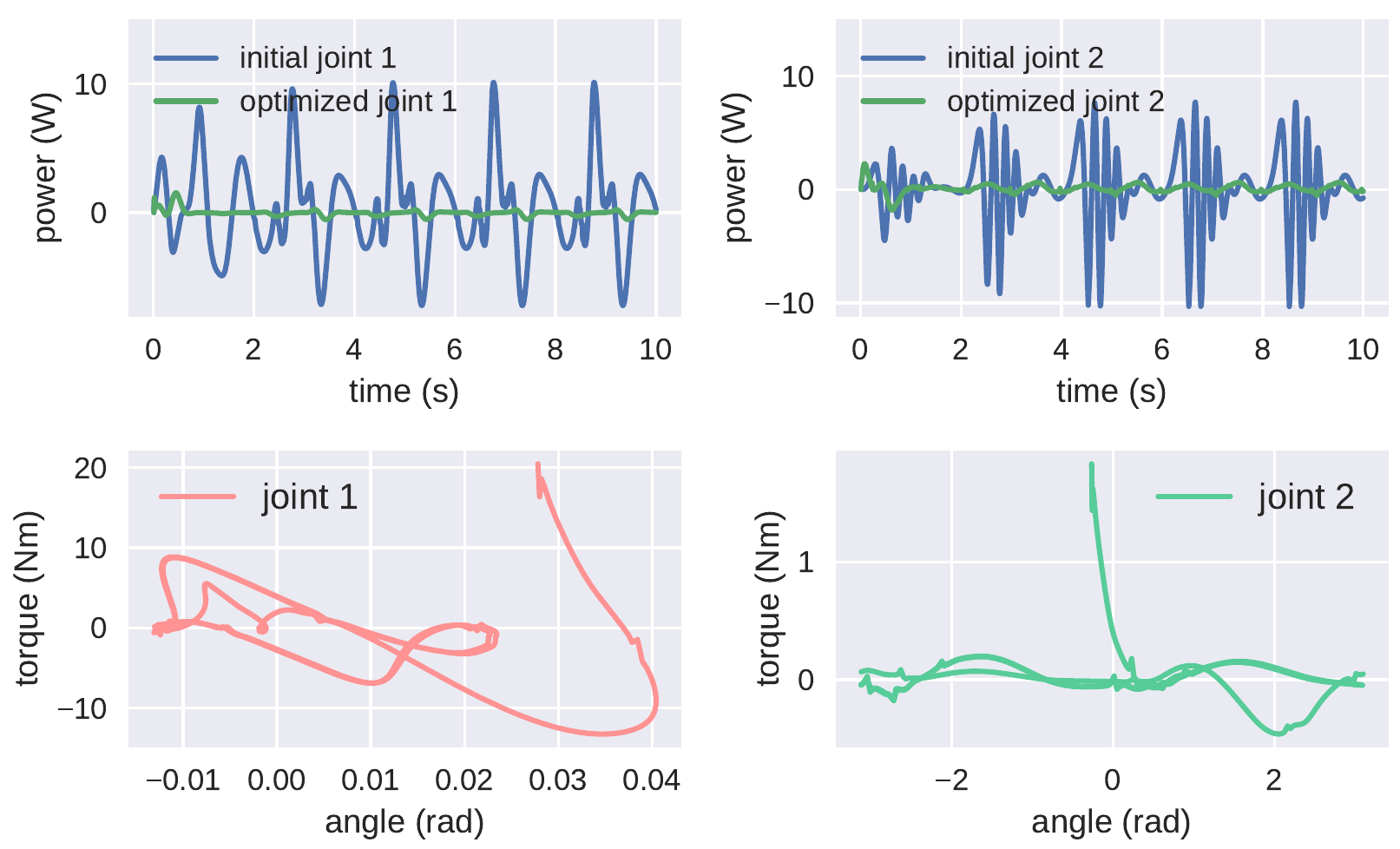}
\caption{The upper figure is the power of joint torque overtime for the double pendulum. Lower left and lower right figures are the torque value change over the angles of two joints. }
\label{fig:torquepower}
\vspace{-0.2in}
\end{figure}

\subsection{Double Pendulum Robot Design}

A double pendulum experiment involving parameter identification is described below. Suppose we have a specified trajectory that we wish to follow repeatedly (e.g., an assembly line) and we wish to determine the link lengths of a robot arm that would best carry out this trajectory while minimizing overall energy expenditure, i.e., $\boldsymbol{\alpha}=\{l_1,l_2\}$. The arm uses a stiffness controller~\cite{salisbury1980active} that calculates motor torque values as
\begin{equation}
\boldsymbol\tau = J^{\top}(\boldsymbol{q})(K(\boldsymbol{p}_{des}-\boldsymbol{p}_{ee}) + D(\dot{\boldsymbol{p}}_{des}-\dot{\boldsymbol{p}}_{ee}))
\label{eq:designcontroller}
\end{equation} 
with stiffness $K=60$ and damping $D=6$. $J^{\top}$ is the robot Jacobian mapping of the pendulum robot, $\boldsymbol{p}_{ee}$ is the Cartesian position of the end effector, and $\boldsymbol{p}_{des}$ is a desired trajectory in Catesian space. In this experiment, $\boldsymbol{p}_{des}$ is a circular trajectory with radius 0.3m in Fig. \ref{fig:pendulumplot}. 
The objective function for parameter estimation includes torque energy and projection loss of simulated trajectory into preset trajectory,

\begin{equation}
    \mathcal{L} = \sum_{t} \norm{ \boldsymbol{p}_{ee}^{t} - \boldsymbol{p}_{des}^{t}}+  \lambda (\boldsymbol{\theta}^{t} - \boldsymbol{\theta}^{t-1}) \cdot \boldsymbol{\tau}^{t}.
\end{equation}
where $\boldsymbol{\theta^{t}}$ is the vector of the angle of two joints at time $t$. The scaling factor $\lambda$ is 25.

The final optimized result is shown in Table \ref{tab:optimizedlength}. The optimal result of the first link and the second link are $l_1=2.707m$ and $l_2=0.302m$. Under the optimized configuration, the pendulum can follow the preset trajectory by changing the second joint angle only. Thus, it has minimal movement and the energy consumption is also optimal.
Fig.\ref{fig:pendulumplot} shows the double pendulum experiment. The initial arm design is insufficient to follow the trajectory, while the optimized one completes the task. Fig. \ref{fig:torquepower} shows the application of the model for optimizing for minimum energy usage as a robot design objective.

\begin{table}[htbp]\centering
\setlength{\tabcolsep}{5pt}
\begin{tabular}{|c|c|c|c|c|}
\hline
    & Link 1 (m) & Link 2 (m) & Loss & Work (J)\\
\hline
Initial & 3.0 & 0.1 & 51.716 & 1.475\\
\hline
Optimal & 2.7026 & 0.3074 & 14.448 & 0.368\\
\hline
\multicolumn{5}{l}{~}\\
\end{tabular}
\caption{Links length, loss and maximum power for the initial case and optimized case.}
\label{tab:optimizedlength}
\end{table}

\begin{table}
\centering
\begin{tabular}{|c|c|c|c|}
\hline
Joint& $s_1$&$e_1$&$w_1$\\
\hline
Mean (Nm)& -2.90e+01& -1.08e+01& -1.20e+00\\
\hline
Max (Nm)& -1.16e+01& -3.99e+00& -5.096e-01\\
\hline
Min (Nm)& -3.91e+01& -1.42e+01& -1.56e+00\\
\hline
MAE (Nm)& 6.69e-01& 4.31e-01& 6.98e-01\\
\hline
\end{tabular}
\caption{Mean, maximum, minimum, and means absolute value (MAE) of estimated gravity compensation torque. We show statistic of joint $s_1$, joint $e_1$, joint $w_1$ which are mainly involved in gravity compensation.}
\label{tab:gravcompenstatic}
\end{table}

\begin{figure}
\vspace{2mm}
\centering
\includegraphics[width=0.99\linewidth]{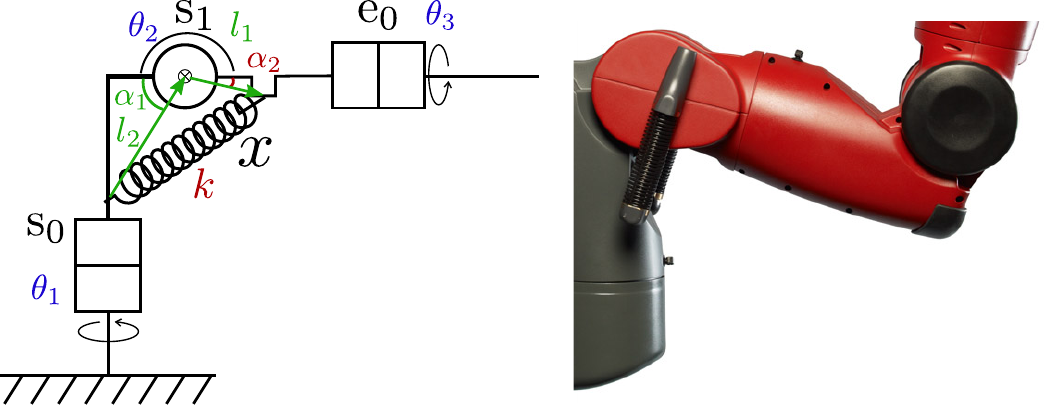}
\caption{Kinematic diagram for joint $s_0$, joint $s_1$, joint $e_0$ and joint $s_1$ spring configuration. Parameters collected from the robot are marked in blue, measured parameters or calculations are marked in green, and parameters for estimation are marked in red.}
\label{fig:JointDiagram}
\vspace{-0.2in}
\end{figure}


\begin{figure}
\vspace{2mm}
\centering
\includegraphics[width=0.99\linewidth]{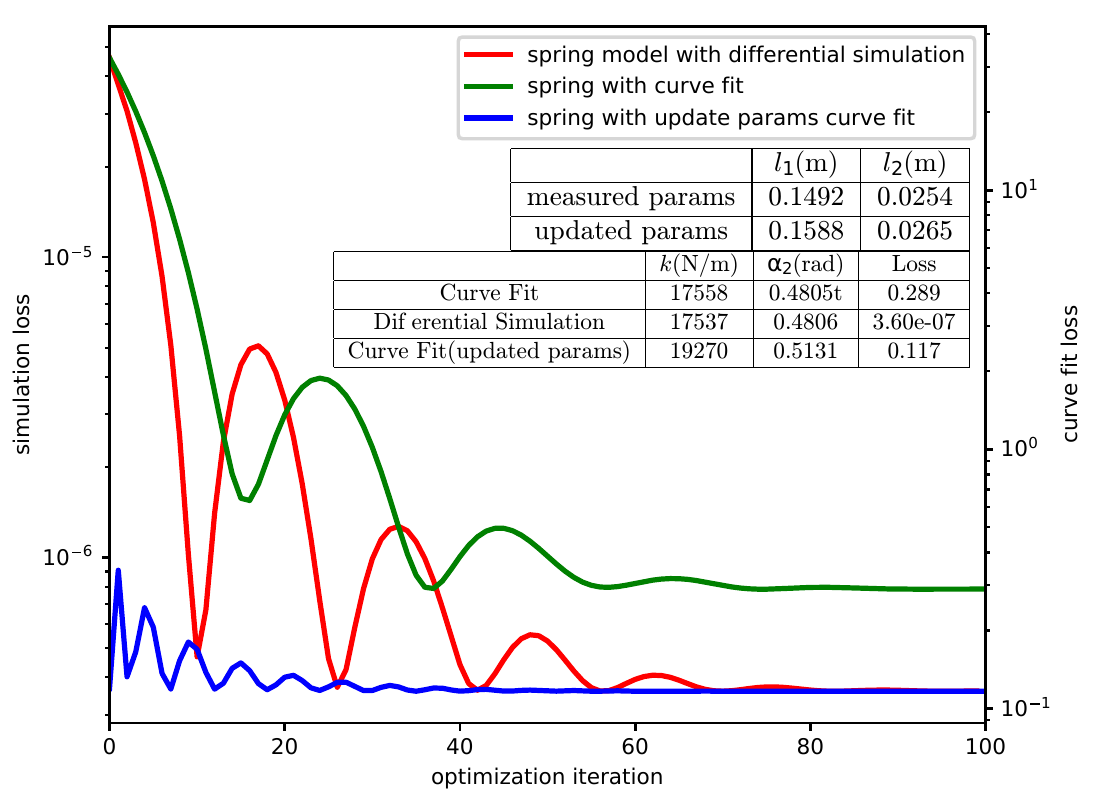}
\caption{Parameter estimation loss during the optimization process. In the upper table, final losses, optimized parameters $k$, $\alpha_2$ are listed for three schemes. For curve fitting, we calculate spring torque with gravity compensation torque and motor torque collected from the robot and perform curve fitting with the spring model and measured or updated parameters. We integrated the spring model into the differentiable simulation method and optimized gravity compensation torques for the differentiable simulation. The upper table shows the value of measured length or updated length.
} 
\label{fig:springloss}
\vspace{-0.2in}
\end{figure}

\subsection{Gravity Compensation and Parameter Estimation}

In this experiment, we consider the Baxter robot from Rethink Robotics that uses 7-DOF series elastic arms, which are not only providing compliance for safety but are additionally set up to provide the arm with a passive gravity counterbalance. One of the challenges with series elastic actuators is that it can be difficult to calibrate as environment conditions, wear and tear, and other factors can often cause expected values to deviate and be non-stationary day to day (thereby affecting gravity counterbalance as well). Thus, we consider the problem of estimating the true spring stiffness $k$ and offset angle $\alpha_2$, i.e., $\boldsymbol{\alpha}=\{k,\alpha_2\}$, for Baxter's shoulder as shown in Figure \ref{fig:JointDiagram}. 

We collect 30 frames of different robot static poses from a real Baxter. Using those pose measurements, we reconstruct the robot in simulation and let the robot stay still for one time step in the simulation. We set $\lambda_{1}=1$, $\lambda_{2}=0.02$ , $\lambda_{3}=0$ for the loss function in Eq. \ref{equ:lossfunctiontorque}. Ground truth torque is then estimated using the Euler-Lagrange equations (Eq. \ref{eq:eulerlagrange}). Table \ref{tab:gravcompenstatic} shows that the mean square error for three joints have similar values, but have less impact on joint $s_1$ and joint $e_1$ than on joint $w_1$ since joint $s_1$ and joint $e_1$ both have relatively large torques values. 
The errors propagate through the kinematic chains and thus a discrepancy between the model and the measurements is seen.

\begin{figure*}
  \begin{minipage}[t]{0.7\linewidth}
    \centering
    \includegraphics[scale=0.5]{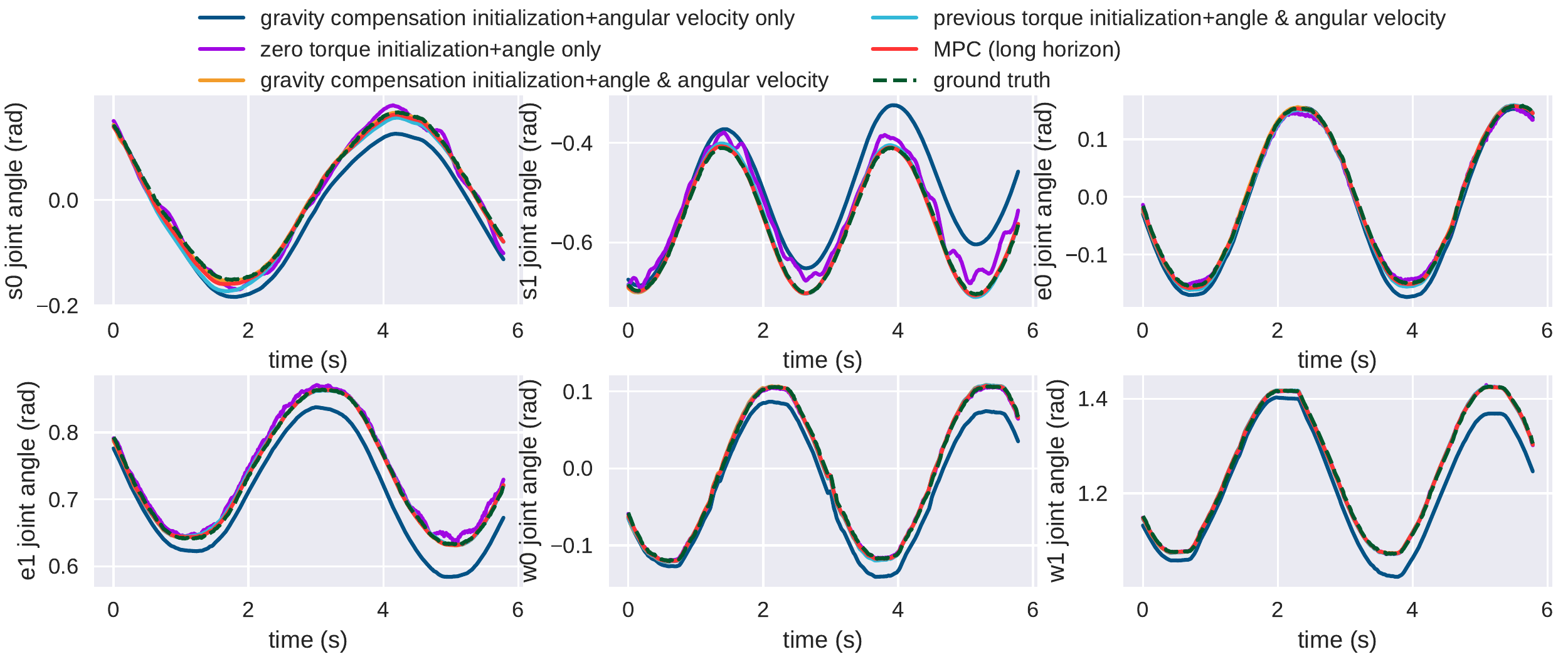}
  \end{minipage}%
  \begin{minipage}[t]{0.4\linewidth}
    \centering
    \includegraphics[scale=0.45]{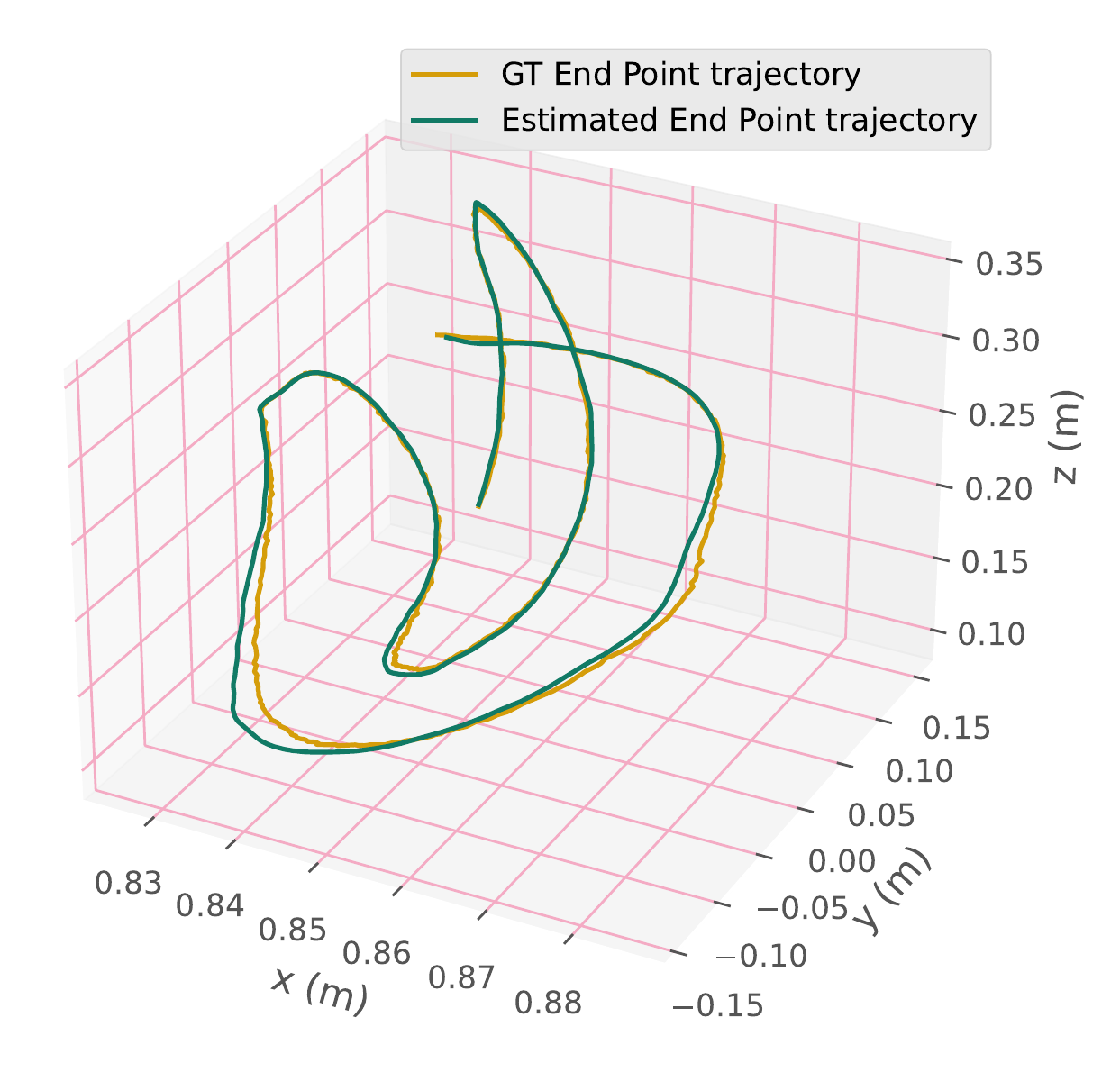}
  \end{minipage}
  
  \caption{\textbf{Left}:Joints angle generated with torque estimation with different     schemes and ground truth trajectory joints angle. From left to right, from up to down, 
    figures show the result for joint $s_0$, joint $s_1$, joint $e_0$, joint $e_1$, joint $w_0$, joint $w_1$ respectively. Zero torque, optimized torque for previous time step or gravity compensation torque are optional initial torque for optimization. Loss function schemes are angle only ($\lambda_{1}=1, \lambda_{2}=0, \lambda_{3}=0$), angular velocity only ($\lambda_{1}=0, \lambda_{2}=1, \lambda_{3}=0$) and angle \& angular velocity ($\lambda_{1}=1, \lambda_{2}=0.02, \lambda_{3}=0$). All schemes use ADAM with 80 or 120 iterations. The learning rate is 1 for zero initialization or 0.2 otherwise.
    \textbf{Right}: Endpoint trajectory visualization of true trajectory and MPC torque estimation simulated trajectory. 
    The MPC has horizon 3 and initializes to previously optimized torque for the first two steps and to gravity compensation torque for the third step. Loss function for MPC implies constraint to limit the torque change rate ($\lambda_{1}=1, \lambda_{2}=0.02, \lambda_{3}=2e-7$). Optimization is performed with an ADAM optimizer with a 0.1 learning rate and 120 iterations.
    }
  \label{fig:anglecompare_endpttraj}
\end{figure*}

The torque provided by the spring on joint $s_1$ is defined as
\begin{equation}
\begin{split}
\tau_{s} &=\sin(\pi-\alpha_1-\alpha_2-\theta_2)\frac{l_1l_2}{x}k(x - x_{rest})\\
x &=\sqrt{2l_{1}l_{2}\cos{(\pi - \alpha_{1} - \alpha_{2})} + l_{1}^{2} + l_{2}^{2}}
\end{split}
\label{equ:SpringTorqueCal}
\end{equation}
where  $l_1$, $l_2$, and $\alpha_2$ are measured or computed with trigonometric relationships.
Spring length $x$ is calculated with the law of cosine and the rest length is 0.2m from the hardware manual \cite{2015_HwSpec}. Thus, The spring torque is calculated with Eq. \ref{equ:SpringTorqueCal}. 
Two estimation schemes, curve fitting, and the differentiable simulation method are used.
For curve fitting, we compute the spring torque $\tau_s$ with ground truth motor torque and estimated gravity compensation torque. With computed spring torque $\tau_s$ and ground truth joint angle $\theta_2$, we can perform curve fitting with stiffness $k$ and offset angle $\alpha_2$ by minimizing the mean square error between computed and estimated spring torque. For differentiable simulation, we integrate the spring model into the differentiable simulation and obtain gradient with respect to $k$ and $\alpha_2$. Similar to gravity compensation, we set $\lambda_1=1$, $\lambda_2=0.2$ for loss function in Eq.\ref{equ:lossfunctiontorque} and let robot stay still.


The optimization loss and result are shown in Figure \ref{fig:springloss}. The use of measured links length ($l_1$, $l_2$) and offset angle $\alpha_2$ for the spring model ensures the curve fitting and the differentiable simulation method converged to similar values. Offset angle $\alpha_2$ did not influence the local minimum value but shifted where it appeared. Only stiffness $k$ is used for estimation since it is coupled with rest length $x_{rest}$. Unlike curve fitting where we compute the gravity compensation torque and then perform optimization, differentiable simulation provides an approach that can estimate parameters in one step.

\begin{figure*}[t!]
\vspace{2mm}
\centering
\includegraphics[width=1.0\textwidth]{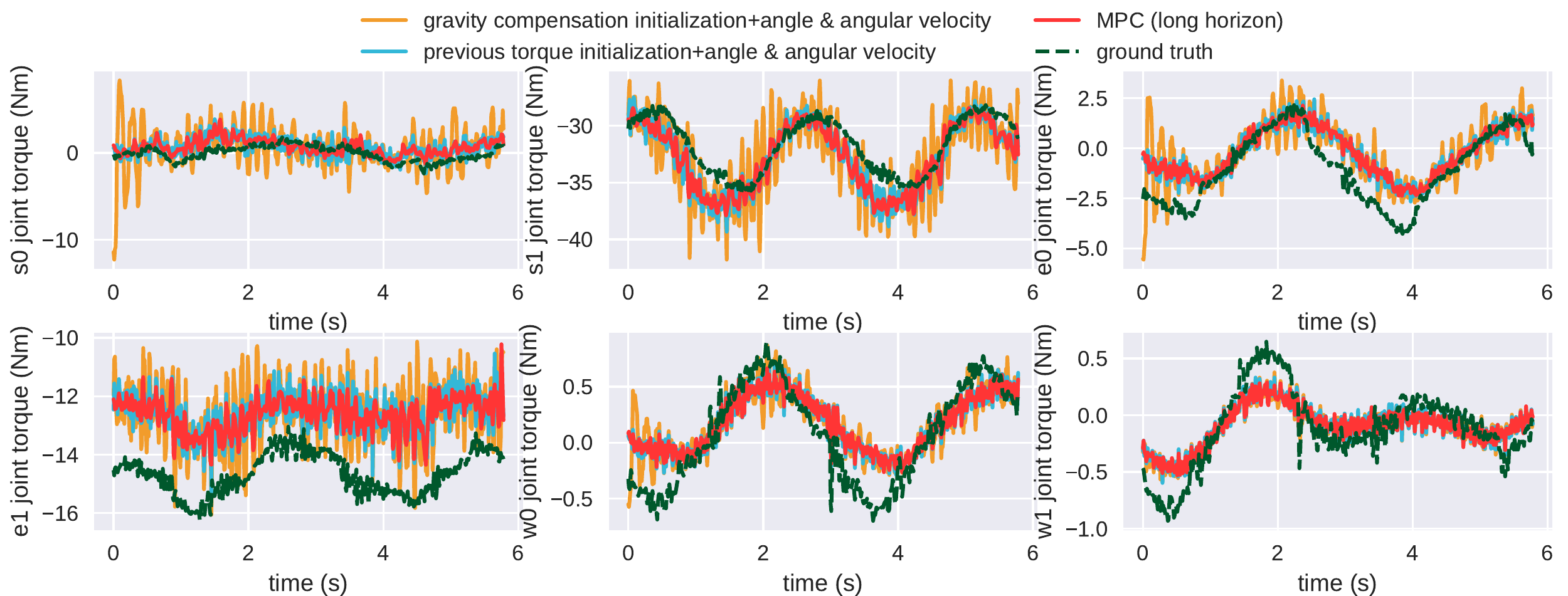}
\caption{Joint torque generated with torque estimation with different schemes and ground truth trajectory joint torques versus time. From left to right, from up to down, figures show the result for joint $s_0$, joint $s_1$, joint $e_0$, joint $e_1$, joint $w_0$, joint $w_1$ respectively. Schemes are specified in Fig. \ref{fig:anglecompare_endpttraj}. 
}
\label{fig:torquecompare}
\end{figure*}

\begin{figure*}[t!]
\vspace{2mm}
\centering
\includegraphics[width=1.0\textwidth]{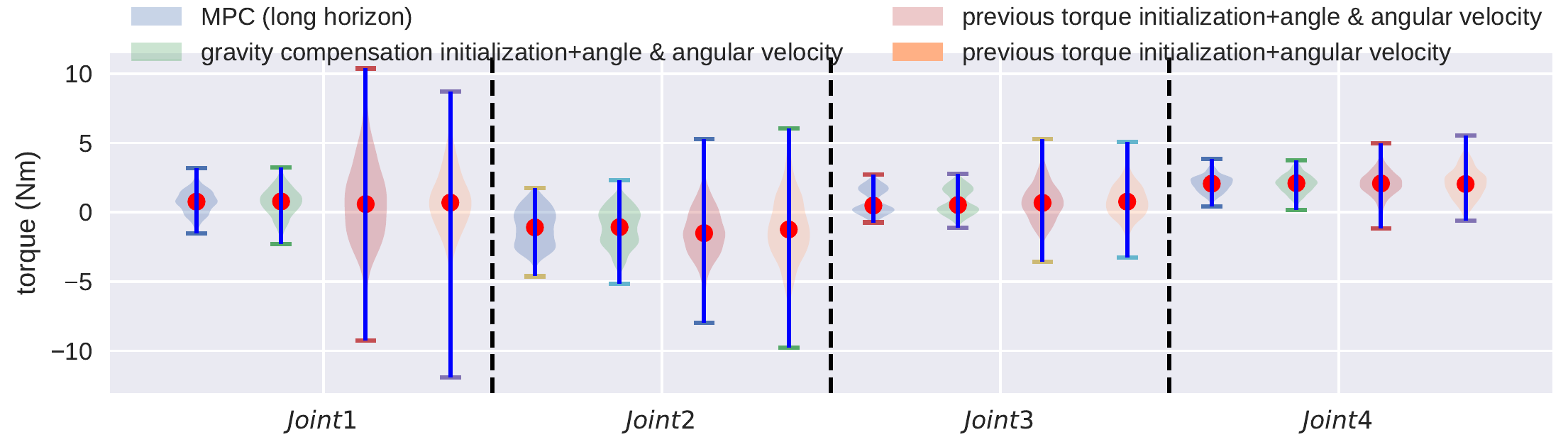}
\caption{Violin plot of torque error with torque estimation schemes for joints $s_0$, joint $s_1$, joint $e_0$, joint $e_1$. 
}
\label{fig:violinplot}
\end{figure*}

We also considered the real possibility that, with the offset angle $\alpha_2$,  the link lengths $l_1$, $l_2$ have some minor deviations from ideal manufacturing specifications. So we estimate link lengths $l_1$, $l_2$ and offset angle $\alpha_2$ independently first, before optimizing stiffness $k$. With these calibrated links length $l_1$ and $l_2$, curve fitting with updated parameters has better performance and the optimized stiffness is closer to reference stiffness compared to without the link length calibration.


\subsection{Real-to-Sim MPC Torque Estimation}

\begin{figure}
\vspace{2mm}
\centering
\includegraphics[width=0.99\linewidth]{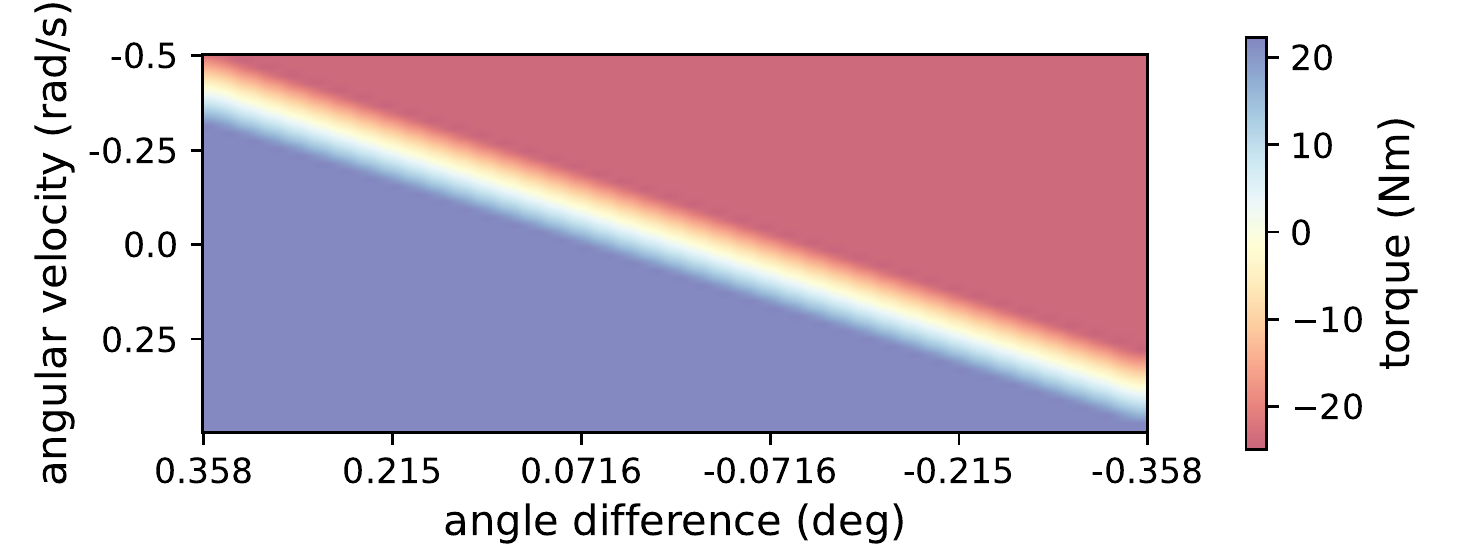}
\caption{Net torque map is a lookup map for $\boldsymbol{\tau}_{net}$ with key theta difference and angular velocity. With the simulation time step being 0.01s, Maximum angular velocity is set to 0.5 rad/s and maximum angular displacement in one-time step is set to 0.00625 rad. Optimized net torque is found for all sampled grid points and interpolated into map space.} 
\label{fig:nettorquemap}
\vspace{-0.2in}
\end{figure}

\begin{table}[]
\setlength{\tabcolsep}{5pt}
\begin{subtable}[t]{0.48\textwidth}
\caption{Root Mean Square Error of joint $s_1$ torque (unit: Nm)}
\begin{tabular}{|c|c|c|c|}
\hline
RMSE & angle only&omega only&angle+omega\\
\hline
zero& 13.429& 2.353 & 31.151\\
\hline
previous torque& 27.269& 2.300& 2.593\\
\hline
gravity compensation& 3.080& 2.424& \textbf{1.7638}\\
\hline
MPC (long horizon)$^{\star}$&  &  & \textbf{1.726}\\
\hline
\end{tabular}
\end{subtable}
\begin{subtable}[t]{0.48\textwidth}
\caption{Root Mean Square Error of joint $e0$ torque (unit: Nm)}
\begin{tabular}{|c|c|c|c|}
\hline
RMSE & angle only&omega only&angle+omega\\
\hline
zero& 2.172 &1.518& 3.293\\
\hline
previous torque& 37.543& 1.500& 1.448\\
\hline
gravity compensation& 1.309& 1.344& \textbf{1.192}\\
\hline
MPC (long horizon)$^{\star}$ &  &  & \textbf{1.196}\\
\hline
\end{tabular}
\end{subtable}

\begin{subtable}[t]{0.48\textwidth}
\caption{Root Mean Square Error of joint $e1$ torque (unit: Nm)}
\begin{tabular}{|c|c|c|c|}
\hline
RMSE & angle only&omega only&angle+omega\\
\hline
zero& 6.926 &2.251& 18.162\\
\hline
previous torque& 32.447 &2.244 &2.309\\
\hline
gravity compensation& 2.745 &2.243 & \textbf{2.164}\\
\hline
MPC (long horizon)$^{\star}$&  &  & \textbf{2.128}\\
\hline
\end{tabular}
\end{subtable}
\caption{Root Mean Square Error of selected joint under three torque initialization schemes, three loss function schemes and MPC schemes. optimal schemes are MPC and gravity compensation initialization with angle \& angular velocity loss. $\star$ indicates that MPC has a different initialization scheme and a different loss function.}
\label{table:rmsetorqueerr}
\end{table}

With Baxter, we tested our method in the context of model-predictive control. The goal is to determine the control inputs to the simulated dynamics model such that the error between a measured trajectory and a modeled trajectory are minimized. This is the real-to-sim problem that has been of more recent interest as a method for imitation and transfer learning\cite{Fei_2021_ICRA,lu2020robust}. MPC is used to determine the control inputs iteratively since solving the full trajectory worth of control inputs at once that results in matching trajectories between sim and real would be difficult. 
A short and long horizon MPC was implemented. For MPC with a short horizon, we optimize the torques for the current time step and use it to update robot states. For MPC with a long horizon, we optimize the torques for 3 time steps simultaneously and update robot states.

To accommodate the use of different sensors (i.e. encoders, IMU) in various circumstances, we design 3 different loss functions for MPC with short horizon: a loss with only joint angles ($\lambda_{1}=1, \lambda_{2}=0, \lambda_{3}=0$), with only angular velocities ($\lambda_{1}=0, \lambda_{2}=1, \lambda_{3}=0$), and with both ($\lambda_{1}=1, \lambda_{2}=0.02, \lambda_{3}=0$).
To start optimization for $\boldsymbol{u}^t$, we choose the torque initialization schemes from zeros, previous optimized torque $\boldsymbol{u}^{t-1}$ and gravity compensation torque.
MPC with long horizon involved a 3-step forward prediction. The torque initialization used the previously optimized torque for the first two steps and the gravity compensation torque from the model for the third step. With multiple steps, we can now assign a value for the loss attributed to preventing rapid torque changes ($\lambda_{1}=1, \lambda_{2}=0.02, \lambda_{3}=2e-7$). 
Fig. \ref{fig:anglecompare_endpttraj} shows our results, demonstrating joint angles nearly match the ground truth for the various configurations of the MPC problem. 
Fig. \ref{fig:torquecompare} and Fig.\ref{fig:violinplot} also show the capability for accurately estimating torque with dynamics and kinematics information.
\begin{figure}[h]
\vspace{2mm}
\centering
\includegraphics[width=0.99\linewidth]{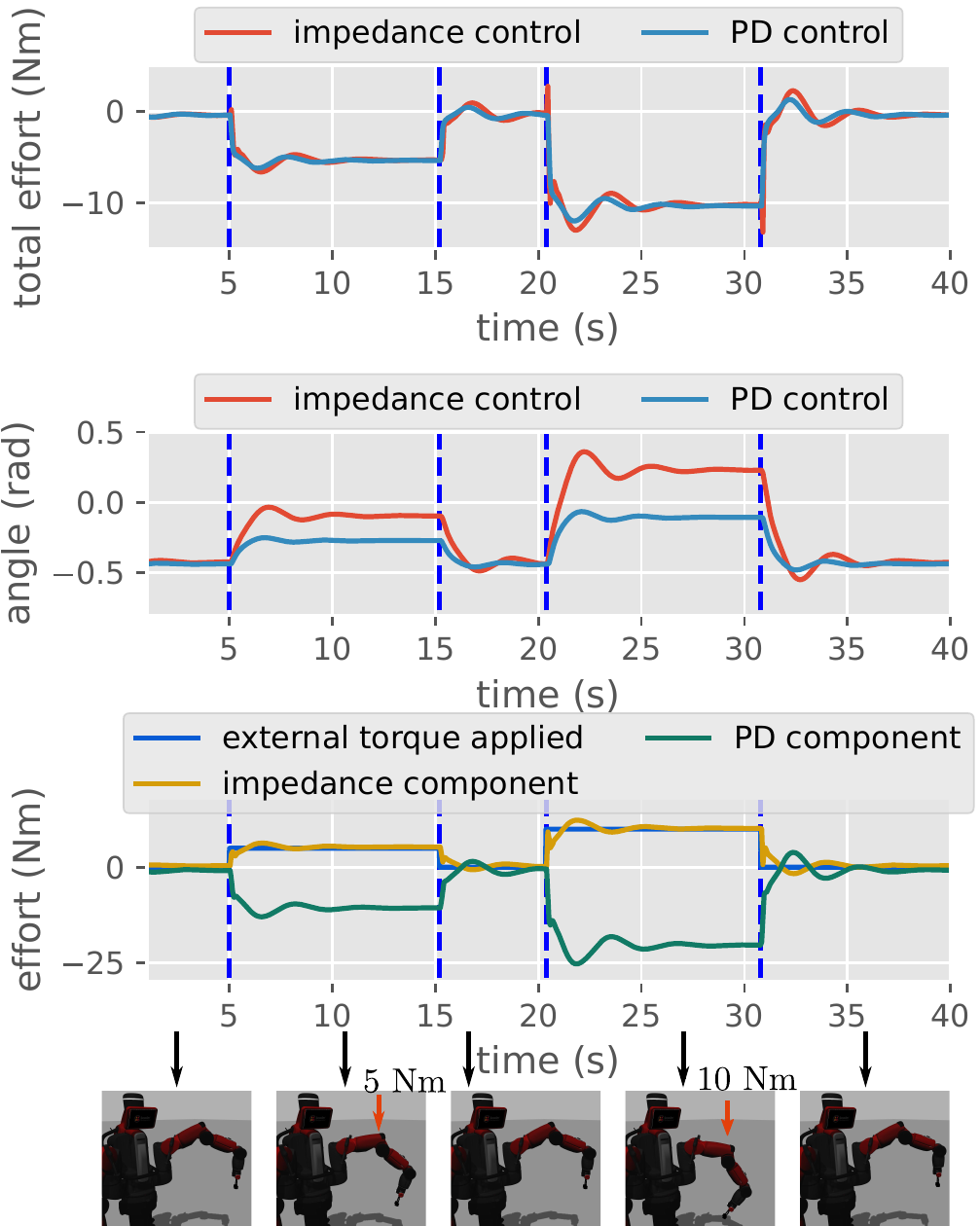}
\caption{Impedance control and PD control for joint $s_1$ in simulation. For both controllers, goal angle $\boldsymbol{q}_d$ is -0.45 rad, goal angular velocity $\dot{\boldsymbol{q}}_d$ is 0 rad/s, and impedance component scaling factor is 1. The external torque of magnitude 5 Nm and 10 Nm are applied at 5s-15s, 20s-30s respectively. The top figure shows the motor torque input for the impedance controller and the PD controller. The middle figure shows the angle change of joint $s_1$ and the bottle figure shows the external torque applied on the joint, impedance component, and PD component in the impedance controller. Note that motor torque in impedance controller is the summation of impedance term multiplied by a scaling factor and PD component. 
} 
\label{fig:simimpedance}
\vspace{-0.2in}
\end{figure}

The RMSE in Table \ref{table:rmsetorqueerr} mainly consists of two parts, the torque offset, the noise when performing torque estimation. In Fig. \ref{fig:torquecompare}, the torque error for joint $e_1$, joint $w_0$, joint $w_1$, occurs on account of unmodel parts of Baxter's arm, and friction noise requires a more significant torque to actuate joints in the real-world than that in simulation. The error also comes from the optimization noise.


\begin{figure}[h]
\vspace{2mm}
\centering
\includegraphics[width=0.99\linewidth]{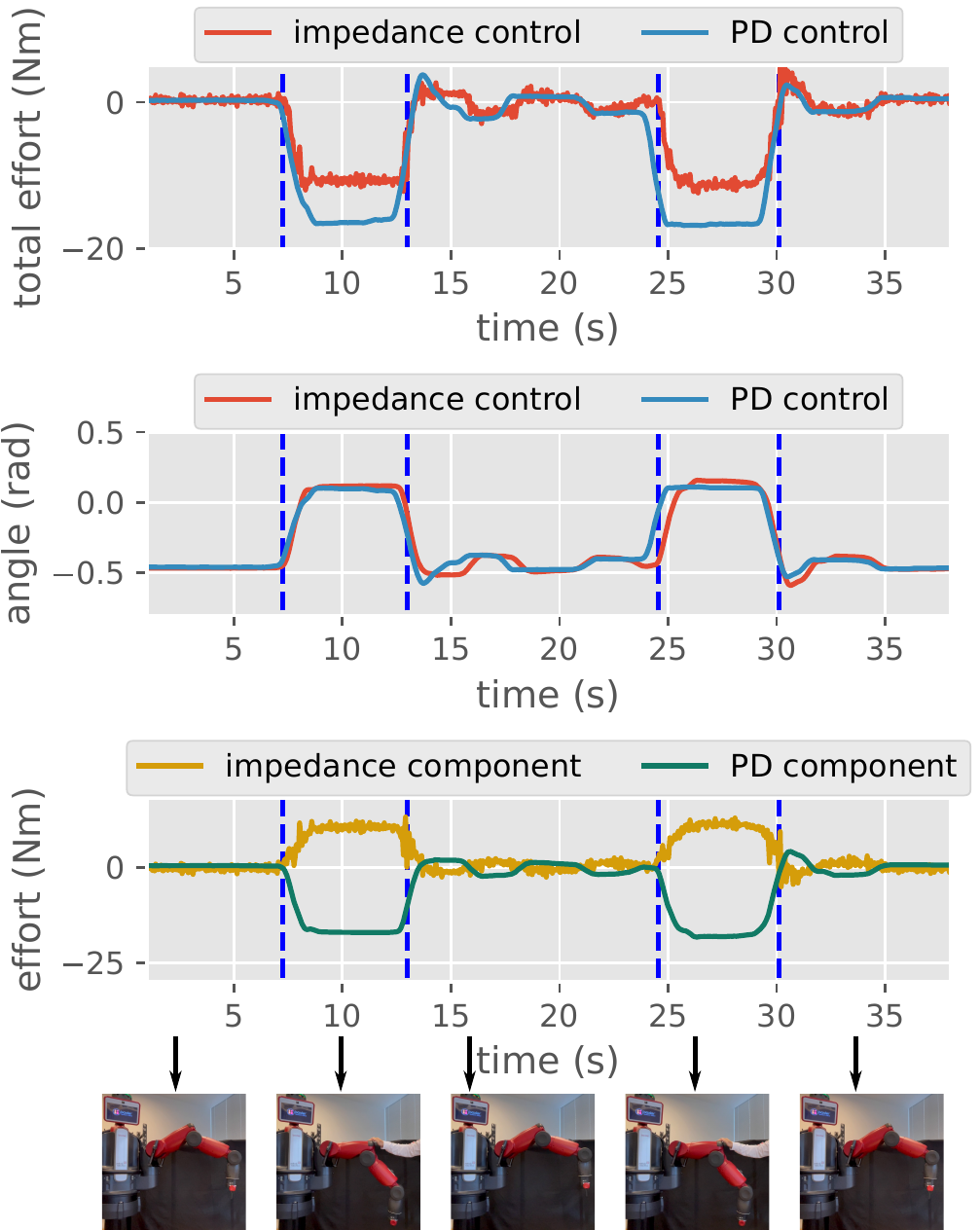}
\caption{Real Baxter robot motor torque and angle of joint $s_1$ when using impedance controller or PD only controller. The goal angle is -0.45 rad and goal angular velocity is 0 rad/s, impedance component scaling factor is 0.6. Manual force is applied to drive joint $s_1$ to about 0 rad. The top figure shows the motor torque input for the impedance controller and PD controller. The middle figure shows the angle change of joint $s_1$, and the bottle figure shows the impedance component and the PD component in the impedance controller.
} 
\label{fig:realimpedance}
\vspace{-0.2in}
\end{figure}

\subsection{Impedance Control}
An impedance control experiment in joint space was conducted for joint $s_1$, following Eq. \ref{eq:impedance_equation} and the use of our simulation framework for calculating various elements within the equation. For simplification, 
the Centrifugal and Coriolis term $\boldsymbol{S} (\boldsymbol{q},\boldsymbol{\dot{q}})\boldsymbol{\dot{q}}$ were not considered; we implemented the PD control term $\tau_{pd}$ and inertia matching term $\boldsymbol{M}(\boldsymbol{q})\boldsymbol{\ddot{q}}_d$. Gravity compensation was handled by Baxter's internal model and is not shown (so only the impedance controller torque is shown). All other joints were fixed during motion. We consider joint $s_1$ angle displacement and angular velocity to optimize net torque. A net torque lookup map, which maps angle difference and angular velocity to net torque $\tau_{net}$ is built with a differentiable simulation framework (Figure \ref{fig:nettorquemap}). To construct such a map, we first sample different pairs of angular velocity and angle differences within the simulator. We set the goal joint angle and initialize angular velocity according to the key's value and optimize to find the control torque (i.e., the net torque). Thus, the loss function for Eq. (\ref{equ:lossfunctiontorque}) only contained angular error ($\lambda_{1}=1, \lambda_{2}=0, \lambda_{3}=0$). 
An external torque is applied twice for 10 seconds in the simulation. Damping and stiffness for the impedance controller is set to $\boldsymbol{D}_m=0.01$  $\boldsymbol{K}_m=30$ respectively. 

Fig. \ref{fig:simimpedance} shows the simulated application of PD control and impedance control in simulation with the torque map as its backbone. 
Fig. \ref{fig:realimpedance} shows the real case,
demonstrating that the contact torque is reduced and impedance control is successfully implemented.


\section{Discussion and Conclusion}
This paper implements a differentiable framework based on position-based dynamics for articulated rigid bodies, with specific application to robot control. With automatic differentiation, we can compute gradients concerning any parameters involved in the simulation and perform gradient-based optimization. This allows one to solve forward and inverse problems that involve position-based dynamics efficiently. Our framework demonstrates its ability in real-to-sim applications, motion control, and parameter identification.
Future work will demonstrate the use of the simulation framework for controlling interactions with more complex environments and objects, leveraging the efficient capability of PBDs for solving contact boundaries down to particle resolution. 



 

\section*{Acknowledgements}
This work was supported by NSF CAREER award $\#$2045803 and the US Army Telemedicine and Advanced Technologies Research Center.

\newpage
\bibliographystyle{ieeetran}
\bibliography{RigidPBD}

\begin{thebibliography}{10}
\providecommand{\url}[1]{#1}
\csname url@rmstyle\endcsname
\providecommand{\newblock}{\relax}
\providecommand{\bibinfo}[2]{#2}
\providecommand\BIBentrySTDinterwordspacing{\spaceskip=0pt\relax}
\providecommand\BIBentryALTinterwordstretchfactor{4}
\providecommand\BIBentryALTinterwordspacing{\spaceskip=\fontdimen2\font plus
\BIBentryALTinterwordstretchfactor\fontdimen3\font minus
  \fontdimen4\font\relax}
\providecommand\BIBforeignlanguage[2]{{%
\expandafter\ifx\csname l@#1\endcsname\relax
\typeout{** WARNING: IEEEtran.bst: No hyphenation pattern has been}%
\typeout{** loaded for the language `#1'. Using the pattern for}%
\typeout{** the default language instead.}%
\else
\language=\csname l@#1\endcsname
\fi
#2}}

\bibitem{Lynch_2017}
K.~M. Lynch and F.~C. Park, \emph{Modern Robotics: Mechanics, Planning, and
  Control}, 1st~ed.\hskip 1em plus 0.5em minus 0.4em\relax USA: Cambridge
  University Press, 2017.

\bibitem{Paszke_2019}
A.~Paszke, S.~Gross, F.~Massa, A.~Lerer, J.~Bradbury, G.~Chanan, T.~Killeen,
  Z.~Lin, N.~Gimelshein, L.~Antiga, A.~Desmaison, A.~Kopf, E.~Yang, Z.~DeVito,
  M.~Raison, A.~Tejani, S.~Chilamkurthy, B.~Steiner, L.~Fang, J.~Bai, and
  S.~Chintala, ``Pytorch: An imperative style, high-performance deep learning
  library,'' in \emph{Advances in Neural Information Processing Systems},
  H.~Wallach, H.~Larochelle, A.~Beygelzimer, F.~d\textquotesingle
  Alch\'{e}-Buc, E.~Fox, and R.~Garnett, Eds., vol.~32.\hskip 1em plus 0.5em
  minus 0.4em\relax Curran Associates, Inc., 2019.

\bibitem{Bradley_2014}
\BIBentryALTinterwordspacing
B.~M. Bell, ``Cppad : A c++ algorithmic differentiation package,'' 2014.
  [Online]. Available: \url{https://coin-or.github.io/CppAD/doc/cppad.htm}
\BIBentrySTDinterwordspacing

\bibitem{todorov_2012}
E.~Todorov, T.~Erez, and Y.~Tassa, ``Mujoco: A physics engine for model-based
  control,'' in \emph{2012 IEEE/RSJ International Conference on Intelligent
  Robots and Systems}.\hskip 1em plus 0.5em minus 0.4em\relax IEEE, 2012, pp.
  5026--5033.

\bibitem{coumans_2021}
E.~Coumans and Y.~Bai, ``Pybullet, a python module for physics simulation for
  games, robotics and machine learning,'' \url{http://pybullet.org},
  2016--2021.

\bibitem{brockman2016openai}
G.~Brockman, V.~Cheung, L.~Pettersson, J.~Schneider, J.~Schulman, J.~Tang, and
  W.~Zaremba, ``Openai gym,'' \emph{arXiv preprint arXiv:1606.01540}, 2016.

\bibitem{Avila_2018}
F.~de~Avila Belbute-Peres, K.~Smith, K.~Allen, J.~Tenenbaum, and J.~Z. Kolter,
  ``End-to-end differentiable physics for learning and control,'' in
  \emph{Advances in Neural Information Processing Systems}, S.~Bengio,
  H.~Wallach, H.~Larochelle, K.~Grauman, N.~Cesa-Bianchi, and R.~Garnett, Eds.,
  vol.~31.\hskip 1em plus 0.5em minus 0.4em\relax Curran Associates, Inc.,
  2018.

\bibitem{Hu_2020DiffTaichi}
\BIBentryALTinterwordspacing
Y.~Hu, L.~Anderson, T.-M. Li, Q.~Sun, N.~Carr, J.~Ragan-Kelley, and F.~Durand,
  ``Difftaichi: Differentiable programming for physical simulation,'' in
  \emph{International Conference on Learning Representations}, 2020. [Online].
  Available: \url{https://openreview.net/forum?id=B1eB5xSFvr}
\BIBentrySTDinterwordspacing

\bibitem{Qiao_2021}
Y.-L. Qiao, J.~Liang, V.~Koltun, and M.~C. Lin, ``Efficient differentiable
  simulation of articulated bodies,'' in \emph{Proceedings of the 38th
  International Conference on Machine Learning}, ser. Proceedings of Machine
  Learning Research, M.~Meila and T.~Zhang, Eds., vol. 139.\hskip 1em plus
  0.5em minus 0.4em\relax PMLR, 18--24 Jul 2021, pp. 8661--8671.

\bibitem{Werling_2021}
K.~Werling, D.~Omens, J.~Lee, I.~Exarchos, and C.~K. Liu, ``Fast and
  feature-complete differentiable physics for articulated rigid bodies with
  contact,'' \emph{arXiv}, 2021.

\bibitem{Heiden_2021neuralsim}
E.~Heiden, D.~Millard, E.~Coumans, Y.~Sheng, and G.~S. Sukhatme, ``Neuralsim:
  Augmenting differentiable simulators with neural networks,'' in \emph{IEEE
  Intl. Conf. on Robotics and Auto. (ICRA)}, 2021.

\bibitem{Tao_Du_2021}
\BIBentryALTinterwordspacing
T.~Du, K.~Wu, P.~Ma, S.~Wah, A.~Spielberg, D.~Rus, and W.~Matusik, ``Diffpd:
  Differentiable projective dynamics,'' \emph{ACM Trans. Graph.}, vol.~41,
  no.~2, nov 2021. [Online]. Available: \url{https://doi.org/10.1145/3490168}
\BIBentrySTDinterwordspacing

\bibitem{Zhiao_Huang_2021}
\BIBentryALTinterwordspacing
Z.~Huang, Y.~Hu, T.~Du, S.~Zhou, H.~Su, J.~B. Tenenbaum, and C.~Gan,
  ``Plasticinelab: {A} soft-body manipulation benchmark with differentiable
  physics,'' \emph{CoRR}, vol. abs/2104.03311, 2021. [Online]. Available:
  \url{https://arxiv.org/abs/2104.03311}
\BIBentrySTDinterwordspacing

\bibitem{Sirui_2021}
\BIBentryALTinterwordspacing
S.~Chen, Y.~Liu, J.~Li, S.~W. Yao, T.~Fan, and J.~Pan, ``Diffsrl: Learning
  dynamic-aware state representation for deformable object control with
  differentiable simulator,'' \emph{CoRR}, vol. abs/2110.12352, 2021. [Online].
  Available: \url{https://arxiv.org/abs/2110.12352}
\BIBentrySTDinterwordspacing

\bibitem{Macklin_2017}
M.~Macklin, M.~Müller, and J.~Bender, ``Position-based simulation methods in
  computer graphics,'' \emph{Eurographics Tutorial}, 2017.

\bibitem{Muller_2020}
M.~M\"{u}ller, M.~Macklin, N.~Chentanez, S.~Jeschke, and T.-Y. Kim, ``Detailed
  rigid body simulation with extended position based dynamics,'' \emph{Computer
  Graphics Forum}, vol.~39, no.~8, pp. 101--112, 2020.

\bibitem{Fei_2021_ICRA}
F.~Liu, Z.~Li, Y.~Han, J.~Lu, F.~Richter, and M.~C. Yip, ``Real-to-sim
  registration of deformable soft tissue with position-based dynamics for
  surgical robot autonomy,'' in \emph{IEEE Intl. Conf. on Robotics and Auto.
  (ICRA)}, 2021.

\bibitem{Yunhai_2021}
Y.~Han, F.~Liu, and M.~C. Yip, ``A 2d surgical simulation framework for
  tool-tissue interaction,'' 2021.

\bibitem{Jingbin_2021_ICRA}
J.~Huang, F.~Liu, F.~Richter, and M.~C. Yip, ``Model-predictive control of
  blood suction for surgical hemostasis using differentiable fluid
  simulations,'' in \emph{IEEE Intl. Conf. on Robotics and Auto. (ICRA)}, 2021.

\bibitem{Schenck_2018}
C.~Schenck and D.~Fox, ``Spnets: Differentiable fluid dynamics for deep neural
  networks,'' in \emph{Proceedings of the Second Conference on Robot Learning
  (CoRL)}, Zurich, Switzerland, 2018.

\bibitem{Jeongseok_2018}
J.~Lee, M.~X. Grey, S.~Ha, T.~Kunz, S.~Jain, Y.~Ye, S.~S. Srinivasa,
  M.~Stilman, and C.~K. Liu, ``Dart: Dynamic animation and robotics toolkit,''
  \emph{Journal of Open Source Software}, vol.~3, no.~22, p. 500, 2018.

\bibitem{Degrave_2019}
J.~Degrave, M.~Hermans, J.~Dambre, and F.~wyffels, ``A differentiable physics
  engine for deep learning in robotics,'' \emph{Frontiers in Neurorobotics},
  vol.~13, p.~6, 2019.

\bibitem{Millard_2020automatic}
D.~Millard, E.~Heiden, S.~Agrawal, and G.~S. Sukhatme, ``Automatic
  differentiation and continuous sensitivity analysis of rigid body dynamics,''
  2020.

\bibitem{Lutter_2020differentiable}
M.~Lutter, J.~Silberbauer, J.~Watson, and J.~Peters, ``Differentiable physics
  models for real-world offline model-based reinforcement learning,'' 2020.

\bibitem{Geilinger_2020}
\BIBentryALTinterwordspacing
M.~Geilinger, D.~Hahn, J.~Zehnder, M.~B\"{a}cher, B.~Thomaszewski, and
  S.~Coros, ``Add: Analytically differentiable dynamics for multi-body systems
  with frictional contact,'' \emph{ACM Trans. Graph.}, vol.~39, no.~6, Nov.
  2020. [Online]. Available: \url{https://doi.org/10.1145/3414685.3417766}
\BIBentrySTDinterwordspacing

\bibitem{Murthy_2021gradsim}
J.~K. Murthy, M.~Macklin, F.~Golemo, V.~Voleti, L.~Petrini, M.~Weiss,
  B.~Considine, J.~Parent-L{\'e}vesque, K.~Xie, K.~Erleben, L.~Paull,
  F.~Shkurti, D.~Nowrouzezahrai, and S.~Fidler, ``gradsim: Differentiable
  simulation for system identification and visuomotor control,'' in
  \emph{International Conference on Learning Representations}, 2021.

\bibitem{Sehoon_2021}
\BIBentryALTinterwordspacing
S.~Ha, S.~Coros, A.~Alspach, J.~Kim, and K.~Yamane, ``Joint optimization of
  robot design and motion parameters using the implicit function theorem,'' in
  \emph{Robotics: Science and Systems XIII, Massachusetts Institute of
  Technology, Cambridge, Massachusetts, USA, July 12-16, 2017}, N.~M. Amato,
  S.~S. Srinivasa, N.~Ayanian, and S.~Kuindersma, Eds., 2017. [Online].
  Available: \url{http://www.roboticsproceedings.org/rss13/p03.html}
\BIBentrySTDinterwordspacing

\bibitem{Xinlei_2021_ICRA}
X.~Pan, A.~Garg, A.~Anandkumar, and Y.~Zhu, ``Emergent hand morphology and
  control from optimizing robust grasps of diverse objects,'' in \emph{2021
  IEEE International Conference on Robotics and Automation (ICRA)}, 2021, pp.
  7540--7547.

\bibitem{Spielberg_2019}
\BIBentryALTinterwordspacing
A.~Spielberg, A.~Zhao, Y.~Hu, T.~Du, W.~Matusik, and D.~Rus,
  ``Learning-in-the-loop optimization: End-to-end control and co-design of soft
  robots through learned deep latent representations,'' in \emph{Advances in
  Neural Information Processing Systems}, H.~Wallach, H.~Larochelle,
  A.~Beygelzimer, F.~d\textquotesingle Alch\'{e}-Buc, E.~Fox, and R.~Garnett,
  Eds., vol.~32.\hskip 1em plus 0.5em minus 0.4em\relax Curran Associates,
  Inc., 2019. [Online]. Available:
  \url{https://proceedings.neurips.cc/paper/2019/file/438124b4c06f3a5caffab2c07863b617-Paper.pdf}
\BIBentrySTDinterwordspacing

\bibitem{Jie_Xu_2021}
\BIBentryALTinterwordspacing
J.~Xu, T.~Chen, L.~Zlokapa, M.~Foshey, W.~Matusik, S.~Sueda, and P.~Agrawal,
  ``An end-to-end differentiable framework for contact-aware robot design,'' in
  \emph{Robotics: Science and Systems}, 2021. [Online]. Available:
  \url{https://doi.org/10.15607/RSS.2021.XVII.008}
\BIBentrySTDinterwordspacing

\bibitem{Pingchuan_2021}
\BIBentryALTinterwordspacing
P.~Ma, T.~Du, J.~Z. Zhang, K.~Wu, A.~Spielberg, R.~K. Katzschmann, and
  W.~Matusik, ``Diffaqua: A differentiable computational design pipeline for
  soft underwater swimmers with shape interpolation,'' \emph{ACM Trans.
  Graph.}, vol.~40, no.~4, jul 2021. [Online]. Available:
  \url{https://doi.org/10.1145/3450626.3459832}
\BIBentrySTDinterwordspacing

\bibitem{Maloisel_2021}
G.~Maloisel, E.~Knoop, C.~Schumacher, and M.~Bächer, ``Automated routing of
  muscle fibers for soft robots,'' \emph{IEEE Transactions on Robotics},
  vol.~37, no.~3, pp. 996--1008, 2021.

\bibitem{Sehee_2019}
\BIBentryALTinterwordspacing
S.~Min, J.~Won, S.~Lee, J.~Park, and J.~Lee, ``Softcon: Simulation and control
  of soft-bodied animals with biomimetic actuators,'' \emph{ACM Trans. Graph.},
  vol.~38, no.~6, nov 2019. [Online]. Available:
  \url{https://doi.org/10.1145/3355089.3356497}
\BIBentrySTDinterwordspacing

\bibitem{Bern_2019}
J.~M. Bern, P.~Banzet, R.~Poranne, and S.~Coros, ``Trajectory optimization for
  cable-driven soft robot locomotion,'' \emph{Robotics: Science and Systems
  XV}, 2019.

\bibitem{Qiao_2021_Multibody}
\BIBentryALTinterwordspacing
Y.-L. Qiao, J.~Liang, V.~Koltun, and M.~Lin, ``Differentiable simulation of
  soft multi-body systems,'' in \emph{Thirty-Fifth Conference on Neural
  Information Processing Systems}, 2021. [Online]. Available:
  \url{https://openreview.net/forum?id=j3fpZLKcXF}
\BIBentrySTDinterwordspacing

\bibitem{Heiden_2021}
E.~Heiden, M.~Macklin, Y.~Narang, D.~Fox, A.~Garg, and F.~Ramos, ``Disect: A
  differentiable simulation engine for autonomous robotic cutting,'' 2021.

\bibitem{Miguel_2021}
\BIBentryALTinterwordspacing
M.~A.~Z. Mora, M.~Peychev, S.~Ha, M.~Vechev, and S.~Coros, ``{\{}PODS{\}}:
  Policy optimization via differentiable simulation,'' 2021. [Online].
  Available: \url{https://openreview.net/forum?id=4f04RAhMUo6}
\BIBentrySTDinterwordspacing

\bibitem{Chiuso_2019}
A.~Chiuso and G.~Pillonetto, ``System identification: A machine learning
  perspective,'' \emph{Annual Review of Control, Robotics, and Autonomous
  Systems}, vol.~2, no.~1, pp. 281--304, 2019.

\bibitem{Ramos_2019}
\BIBentryALTinterwordspacing
F.~Ramos, R.~C. Possas, and D.~Fox, ``Bayessim: adaptive domain randomization
  via probabilistic inference for robotics simulators,'' \emph{CoRR}, vol.
  abs/1906.01728, 2019. [Online]. Available:
  \url{http://arxiv.org/abs/1906.01728}
\BIBentrySTDinterwordspacing

\bibitem{Lutter_2021}
\BIBentryALTinterwordspacing
M.~Lutter, J.~Silberbauer, J.~Watson, and J.~Peters, ``A differentiable
  newton-euler algorithm for real-world robotics,'' \emph{CoRR}, vol.
  abs/2110.12422, 2021. [Online]. Available:
  \url{https://arxiv.org/abs/2110.12422}
\BIBentrySTDinterwordspacing

\bibitem{Lidec_2021}
Q.~Le~Lidec, I.~Kalevatykh, I.~Laptev, C.~Schmid, and J.~Carpentier,
  ``Differentiable simulation for physical system identification,'' \emph{IEEE
  Robotics and Automation Letters}, vol.~6, no.~2, pp. 3413--3420, 2021.

\bibitem{Wang_Kun_2020_pmlr}
\BIBentryALTinterwordspacing
K.~Wang, M.~Aanjaneya, and K.~Bekris, ``A first principles approach for
  data-efficient system identification of spring-rod systems via differentiable
  physics engines,'' in \emph{Proceedings of the 2nd Conference on Learning for
  Dynamics and Control}, ser. Proceedings of Machine Learning Research, A.~M.
  Bayen, A.~Jadbabaie, G.~Pappas, P.~A. Parrilo, B.~Recht, C.~Tomlin, and
  M.~Zeilinger, Eds., vol. 120.\hskip 1em plus 0.5em minus 0.4em\relax PMLR,
  10--11 Jun 2020, pp. 651--665. [Online]. Available:
  \url{https://proceedings.mlr.press/v120/wang20b.html}
\BIBentrySTDinterwordspacing

\bibitem{Carolyn_2020}
C.~Matl, Y.~Narang, R.~Bajcsy, F.~Ramos, and D.~Fox, ``Inferring the material
  properties of granular media for robotic tasks,'' in \emph{2020 IEEE
  International Conference on Robotics and Automation (ICRA)}, 2020, pp.
  2770--2777.

\bibitem{Jiajun_2015}
\BIBentryALTinterwordspacing
J.~Wu, I.~Yildirim, J.~J. Lim, B.~Freeman, and J.~Tenenbaum, ``Galileo:
  Perceiving physical object properties by integrating a physics engine with
  deep learning,'' in \emph{Advances in Neural Information Processing Systems},
  C.~Cortes, N.~Lawrence, D.~Lee, M.~Sugiyama, and R.~Garnett, Eds.,
  vol.~28.\hskip 1em plus 0.5em minus 0.4em\relax Curran Associates, Inc.,
  2015. [Online]. Available:
  \url{https://proceedings.neurips.cc/paper/2015/file/d09bf41544a3365a46c9077ebb5e35c3-Paper.pdf}
\BIBentrySTDinterwordspacing

\bibitem{Muller_2005_ShapeMatching}
\BIBentryALTinterwordspacing
M.~M\"{u}ller, B.~Heidelberger, M.~Teschner, and M.~Gross, ``Meshless
  deformations based on shape matching,'' \emph{ACM Trans. Graph.}, vol.~24,
  no.~3, p. 471–478, jul 2005. [Online]. Available:
  \url{https://doi.org/10.1145/1073204.1073216}
\BIBentrySTDinterwordspacing

\bibitem{Muller_2011_OrientedParticle}
\BIBentryALTinterwordspacing
M.~M\"{u}ller and N.~Chentanez, ``Solid simulation with oriented particles,''
  \emph{ACM Trans. Graph.}, vol.~30, no.~4, jul 2011. [Online]. Available:
  \url{https://doi.org/10.1145/2010324.1964987}
\BIBentrySTDinterwordspacing

\bibitem{Deul_2016}
\BIBentryALTinterwordspacing
C.~Deul, P.~Charrier, and J.~Bender, ``Position-based rigid-body dynamics,''
  \emph{Comput. Animat. Virtual Worlds}, vol.~27, no.~2, p. 103–112, mar
  2016. [Online]. Available: \url{https://doi.org/10.1002/cav.1614}
\BIBentrySTDinterwordspacing

\bibitem{Sola_2017quaternion}
J.~Solà, ``Quaternion kinematics for the error-state kalman filter,'' 2017.

\bibitem{PyTorch}
``Pytorch,'' \url{https://pytorch.org/}, accessed: 2021-11.

\bibitem{Lutter_2020}
\BIBentryALTinterwordspacing
M.~Lutter, J.~Silberbauer, J.~Watson, and J.~Peters, ``Differentiable physics
  models for real-world offline model-based reinforcement learning,''
  \emph{CoRR}, vol. abs/2011.01734, 2020. [Online]. Available:
  \url{https://arxiv.org/abs/2011.01734}
\BIBentrySTDinterwordspacing

\bibitem{salisbury1980active}
J.~K. Salisbury, ``Active stiffness control of a manipulator in cartesian
  coordinates,'' in \emph{1980 19th IEEE conference on decision and control
  including the symposium on adaptive processes}.\hskip 1em plus 0.5em minus
  0.4em\relax IEEE, 1980, pp. 95--100.

\bibitem{2015_HwSpec}
\BIBentryALTinterwordspacing
{Rethink Robotics.} (2015) Baxter research robot hardware specifications.
  [Online]. Available:
  \url{https://sdk.rethinkrobotics.com/wiki/Hardware_Specifications}
\BIBentrySTDinterwordspacing

\bibitem{lu2020robust}
J.~Lu, F.~Richter, and M.~Yip, ``Robust keypoint detection and pose estimation
  of robot manipulators with self-occlusions via sim-to-real transfer,''
  \emph{arXiv preprint arXiv:2010.08054}, 2020.

\end{thebibliography}

\end{document}